\documentclass[11pt]{article}

\usepackage[preprint]{acl}
\usepackage{enumitem}
\usepackage{times}
\usepackage{latexsym}

\usepackage[T1]{fontenc}

\usepackage[utf8]{inputenc}
\usepackage{amsmath} 
\usepackage{microtype}

\usepackage{inconsolata}

\usepackage{graphicx}

\usepackage{float}
\usepackage{ragged2e}
\usepackage{amsmath,amsfonts}
\usepackage{algorithmic}
\usepackage{graphicx}
\usepackage{enumerate}
\usepackage{pifont,tabularx,booktabs}
\usepackage{lscape}
\usepackage{pifont}
\usepackage{booktabs} 
\usepackage{makecell} 
\usepackage{enumitem}
\usepackage{array}
\usepackage{caption} 
\usepackage{siunitx} 
\usepackage{tabularx}
\usepackage{bm}
\usepackage[framemethod=TikZ]{mdframed}
\usepackage[dvipsnames,svgnames,table]{xcolor}
\usepackage{mdframed} 
\usepackage{textcomp}
\usepackage{url}
\usepackage{enumitem}
\usepackage{amsthm}
\usepackage{multirow}
\usepackage{natbib} 
\usepackage{changepage}
\theoremstyle{plain}

\newcommand{\highlightcell}{\cellcolor{blue!12}}

\usepackage{url}
\newcommand{\cmark}{\ding{51}}   
\newcommand{\xmark}{\ding{55}}   

\newcommand{\promptbox}[3]{%
  \begin{mdframed}[
      linewidth=1pt,
      linecolor=black,
      backgroundcolor=blue!5, 
      roundcorner=4pt,
      innertopmargin=6pt, innerbottommargin=6pt]
    #2
  \end{mdframed}
  \begin{center}\small
    Prompt\,#1: #3
  \end{center}\vspace{4pt}
}
\title{Assessing Capabilities of Large Language Models in Social Media Analytics: A Multi-task Quest}

\author{
Ramtin Davoudi \\
Utah State University \\
\texttt{ramtin.davoudi@usu.edu}
\And
Kartik Thakkar \\
Utah State University \\
\texttt{kartik.thakkar@usu.edu}
\And
Nazanin Donyapour \\
Independent Researcher \\
\texttt{nazanin.donyapour@gmail.com}
\AND
Tyler Derr \\
Vanderbilt University \\
\texttt{tyler.derr@vanderbilt.edu}
\And
Hamid Karimi \\
Utah State University \\
\texttt{hamid.karimi@usu.edu}
}

\begin{document}
\maketitle
\begin{abstract}

In this study, we present the first comprehensive evaluation of modern LLMs—including GPT-4, GPT-4o, GPT-3.5-Turbo, Gemini 1.5 Pro, DeepSeek-V3, Llama 3.2, and BERT—across three core social media analytics tasks on a Twitter (X) dataset: (I) Social Media Authorship Verification, (II) Social Media Post Generation, and (III) User Attribute Inference. For the authorship verification, we introduce a systematic sampling framework over diverse user and post selection strategies and evaluate generalization on newly collected tweets from January 2024 onward to mitigate “seen-data” bias. For post generation, we assess the ability of LLMs to produce authentic, user-like content using comprehensive evaluation metrics. Bridging Tasks I and II, we conduct a user study to measure real users’ perceptions of LLM-generated posts conditioned on their own writing. For attribute inference, we annotate occupations and interests using two standardized taxonomies (IAB Tech Lab 2023 and 2018 U.S. SOC) and benchmark LLMs against existing baselines. Overall, our unified evaluation provides new insights and establishes reproducible benchmarks for LLM-driven social media analytics. The code and data are provided in the supplementary material and will also be made publicly available upon publication.
\end{abstract}

\section{Introduction}
\label{sec:introduction}

Online social media platforms have become integral to modern society, generating vast amounts of user-generated content that offers unique insights into various domains such as marketing~\cite{singh2018impact}, public health~\cite{schillinger2020infodemics}, crisis management~\cite{saroj2020use}, and so on. This has led to a unique and vibrant research field known as \textit{social media analytics} (or social media mining). While there has been significant progress in social media analytics since the emergence of popular social networking platforms such as Facebook and Twitter (now X), it still faces critical challenges in fully leveraging the insights and potential of social media data. One of the main challenges is the complexity of user-generated content, especially the text. The social media content is often informal, ambiguous, or irrelevant (e.g., spam and memes), making content analysis complex. Also, social media language changes rapidly (e.g., new slang, memes), so models trained on past data may quickly become outdated. 


Recent advances in large language models (LLMs), including GPT, DeepSeek, and Gemini, offer new opportunities to address these challenges. Trained on large and diverse corpora, LLMs demonstrate strong capabilities in understanding and generating language in dynamic and noisy environments. They have shown competitive performance across tasks such as sentiment analysis~\cite{zhang2023sentiment}, stance detection~\cite{gambini2024evaluating}, misinformation classification~\cite{hu2024bad}, topic extraction~\cite{mu2024large}, and summarization~\cite{zhang2024benchmarking}, often with minimal task-specific supervision.

In this work, we conduct a comprehensive empirical evaluation of LLMs across three core social media analytics tasks: \textbf{(I) Social Media Authorship Verification}, \textbf{(II) Social Media Post Generation}, and \textbf{(III) User Attribute Inference}. Task I aims to determine whether a post was authored by a specific user. While authorship verification has applications in digital forensics and plagiarism detection~\cite{tyo2022state}, social media settings pose additional challenges due to short, low-quality content~\cite{huang2025authorship}. Task II focuses on generating posts that reflect a user’s style and content preferences. Although automated text generation has been widely studied~\cite{celikyilmaz2020evaluation}, producing realistic and personalized social media content remains difficult~\cite{perez2023efficiency,du2023automatic}, and LLMs are increasingly being leveraged for personalized content generation~\cite{kumar2024longlamp,salemi2024lamp,zhang2024personalization}. 
Task III addresses the prediction of user attributes (e.g., occupation, interests) from textual content. Such inference supports personalization and large-scale behavioral analysis~\cite{hu2007demographic,de2010predicting,goel2014predicting}, yet remains challenging due to label ambiguity and fine-grained classification requirements. For a detailed discussion of work related to these three tasks, we refer readers to Appendix~\ref{sec:appendix_G}.


\textbf{Why these three tasks together?} Social media characterizes \textit{user identity} through observable traces in user-generated content~\cite{gunduz2017effect,shulman2022self}. Similar to previous studies, we conceptualize this identity at three levels: the user's unique identity through their social media posts (Task I)~\cite{yadav2017social}, the user's unique social media signature, style, and tone (Task II)~\cite{abbasi2008writeprints}, and latent demographic attributes (Task III)~\cite{tigunova2020reddust}. Thus, together, these tasks enable a coherent assessment of how LLMs infer and recognize user identity.

In this paper, we evaluate several state-of-the-art LLMs—including GPT, Gemini, DeepSeek, Llama, and BERT—alongside traditional ML baselines on a large Twitter (X) dataset. Our contributions are: 

\begin{itemize}[topsep=0pt, partopsep=0pt, parsep=0pt, itemsep=0pt, leftmargin=16pt]
    \item To the best of our knowledge, this is the first study to jointly and systematically assess multiple modern LLMs across three fundamental social media analytics tasks under a unified evaluation framework.

    \item For \textbf{Social Media Authorship Verification} (Task I), we design diverse user and post \textbf{sampling strategies} to enable controlled and realistic benchmarking. We further address ``\textbf{seen data}'' (data leakage) bias by evaluating models on newly collected tweets from January 2024 onward, testing generalization beyond training cut-off periods.

    \item For \textbf{Social Media Post Generation} (Task II), we introduce a structured evaluation protocol for user-conditioned generation based on a curated set of verified, active, and profile-rich users, combining lexical, semantic, and diversity-based metrics to characterize trade-offs between authenticity and fluency.

    \item Bridging Tasks I and II, we conduct the \textbf{first user study} measuring real users’ perceptions of authenticity in LLM-generated tweets, enabling direct comparison between automatic metrics and human judgments.

    \item For \textbf{User Attribute Inference} (Task III), we predict users’ \textit{occupations} and \textit{interests} using two standardized taxonomies: the IAB Tech Lab Content Taxonomy v3.1 and the 2018 U.S. Standard Occupational Classification (SOC). Grounding evaluation in these multi-level ontologies enables reproducible labeling, hierarchical analysis from coarse to fine granularity, and alignment with established real-world classification standards.
\end{itemize}

Collectively, this work establishes a unified and reproducible benchmarking framework for analyzing the capabilities and limitations of modern LLMs in social media analytics.

\textbf{Remark:} In this study, we use Gemini 1.5 Pro, DeepSeek-V3, Llama 3.2, and three versions of GPT (4, 4o, and 3.5-Turbo). For brevity, we drop the versions from Gemini, DeepSeek, and Llama. 

\section{Methodology}
\label{sec:methodology}

\begin{figure*}[h]
    \centering
    \includegraphics[width=0.90\linewidth]{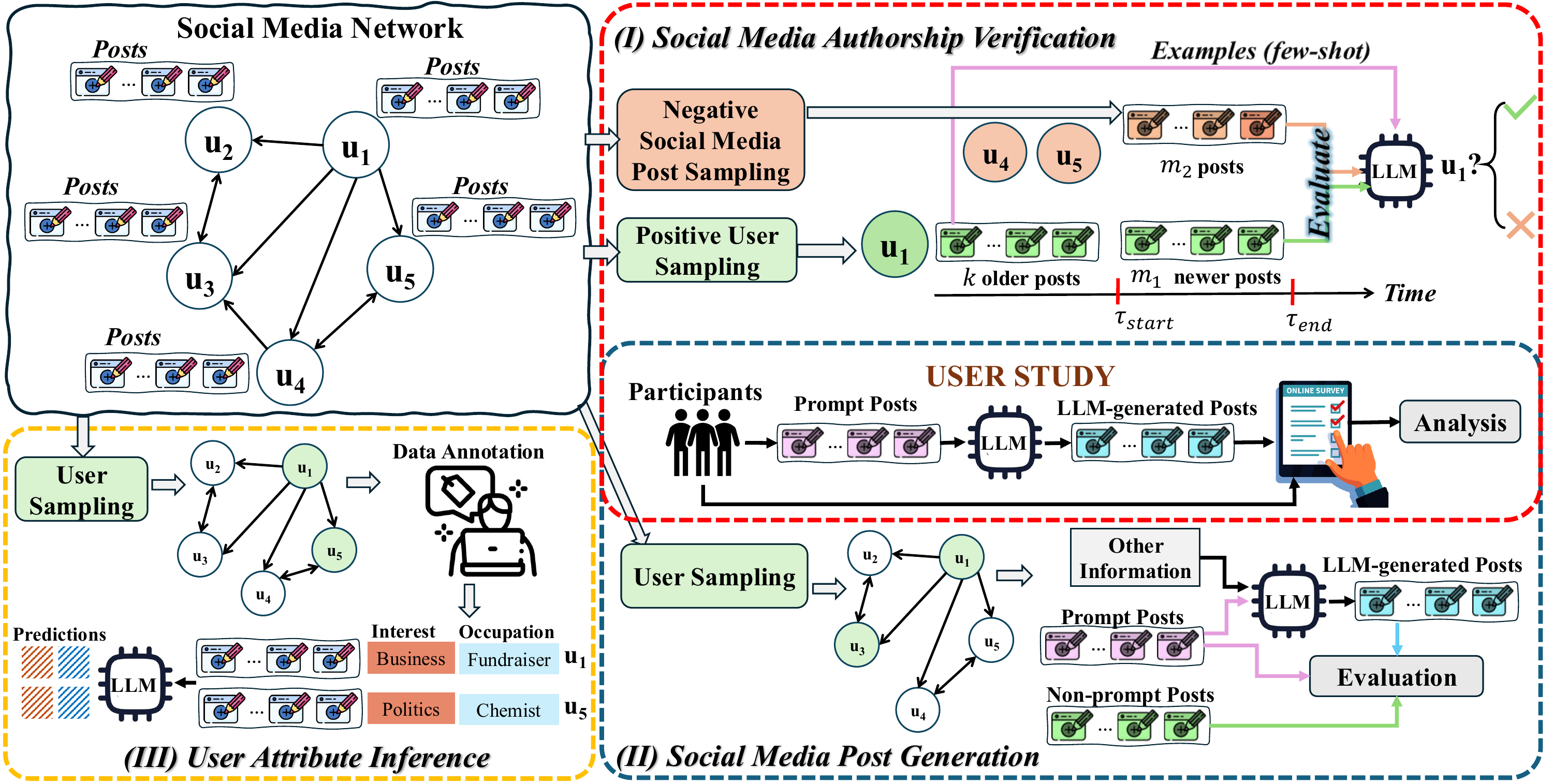}
    \caption{An overview of the proposed methodology across three social media analytic tasks  }
    \label{fig:overall}
    \vskip -1ex
\end{figure*}

Figure~\ref{fig:overall} demonstrates an overview of the proposed methodology for investigating the capabilities of LLMs across three social media analytic tasks.  All three tasks are fed from a social media network dataset consisting of the temporal friendship graph and the user-generated posts acquired from~\cite{kheiri2023analysis} (see the Dataset in Appendix~\ref{sec:appendix_A}  for more information). {We focus on Twitter (X) due to its text-centric nature and well-defined relational signals (e.g., follower networks and various interactions), which make it suitable for controlled benchmarking. The dataset includes temporal network structure and user-generated posts, enabling evaluation of authorship verification, post generation, and attribute inference. Although our experiments use Twitter data, the methodology is platform-agnostic and can be adapted to other platforms by redefining relational and interaction signals.  Next, we explain each component of this method.

\subsection{Social Media Authorship Verification}
\label{subsec:authorship_verification}

In this task, we frame a binary classification task in which an LLM must distinguish between social media posts genuinely authored by a target user and those authored by other users. Next, we explain each part of this task.

\noindent \textbf{Positive User (and Post) Sampling.} Positive (or target) users are users whose posts are assessed by an LLM. The purpose of this sampling is twofold: (1) The sheer high number of social media users in our dataset (around 120K) makes it impossible to assess all of the users (mainly due to the cost); and (2) We can sample different users to investigate LLMs' power across diverse groups. We sample three sets of target (positive) users:
\begin{itemize}[topsep=0pt, partopsep=0pt, parsep=0pt, itemsep=0pt, leftmargin=16pt]
    \item \textbf{Random (Rnd)}: A simple random selection of users.
    \item \textbf{Recent Active (Rec)}: Users active within the last three weeks in our dataset.
    \item \textbf{Top Active Users (Top)}: Users with the highest overall number of posts (tweets).
\end{itemize}

Once a positive user is selected (shown in green color in Figure~\ref{fig:overall}), we chronologically sort their posts (posts shown in green in Figure~\ref{fig:overall}). Then, we sample two sets of posts: (1) \textit{Few-shot Example Posts}: $k$ older posts and (2) \textit{Positive Evaluation Posts}: $m_1$ newer posts--See Figure~\ref{fig:overall}. 
The former is used as training examples for LLM to "become familiar" with the positive user's content (i.e., employing few-shot samples), while the latter is used to form the binary classification described below.        








\textbf{Negative Social Media Post Sampling.} To form the binary classification task for an LLM, for each positive user, we also sample $m_2$ posts from other users called \textit{Negative Evaluation Posts} (shown in orange color in Figure~\ref{fig:overall}). To make a fair comparison, we sample these posts at the same time interval as the \textit{Positive Evaluation Posts}. More specifically, let the target user’s \(m_1\) \textit{Positive Evaluation Posts} correspond to a time interval \(\bigl[\tau_{\text{start}},\,\tau_{\text{end}}\bigr]\), where \(\tau_{\text{start}}\) and \(\tau_{\text{end}}\) are the earliest and latest timestamps of these posts, respectively (see Figure~\ref{fig:overall}). Then, we employ one of the following strategies for \textit{Negative Evaluation Posts} within the same time interval of \(\bigl[\tau_{\text{start}},\,\tau_{\text{end}}\bigr]\):

\begin{itemize}[topsep=0pt, partopsep=0pt, parsep=0pt, itemsep=0pt, leftmargin=16pt]
    \item \textbf{Random Sampling}: A simple random draw of $m_2$ posts from a pool of other users’ posts.

    \item \textbf{Similar Topics Sampling}: 
    We embed each of the \(m_1\) \textit{Positive Evaluation Posts} into vector representations (e.g., using a text embedding model). Similarly, we embed candidate tweets from other users. Then, we calculate each candidate post’s average similarity to the user’s \textit{Positive Evaluation Posts} and select the top $m_2$ most similar posts to form  \textit{Negative Evaluation Posts}. We call this sampling strategy \textit{Topic-similar}.

    \item \textbf{Social Graph-based Sampling}: 
    We leverage the social graph to draw \textit{Negative Evaluation Posts}. For each positive user $u_i$, we randomly select $m_2$ posts from other users who are \textit{followees} ($\{\forall v | v \leftarrow u_i\}$), \textit{followers} ($\{\forall v | v \rightarrow u_i\}$), or \textit{Reciprocal} ($\{\forall v | v \leftrightarrow u_i\}$) of $u_i$, making three sample sets.  This sampling results in three sets, referred to as  \textit{Followers-only}, \textit{Followees-only}, and \textit{Reciprocal}.  
    
\end{itemize}



\noindent \textbf{LLM's Prompt.} After selecting a target (positive) user and their $k$ \textit{Few-shot Example Posts} and forming both \(m_1\) \textit{Positive Evaluation Posts}, \(m_2\)  \textit{Negative Evaluation Posts}, we present each classification instance to the LLM. The exact prompt is in Appendix~\ref{sec:appendix_B} (Prompt 1).

\noindent \textbf{Unbiased (Unseen) Data Investigation.} To evaluate authorship verification without bias from potentially memorized data, we also examine the issue of ``seen data'' by using social media posts (tweets) posted from  Jan. 2024 onward--well beyond the LLMs' training cutoff dates (see Appendix~\ref{sec:appendix_C}, Table~\ref{tab:model_cutoffs} for exact cutoff dates). Specifically, from an initial pool of systematically filtered active users (\texttt{VAPOR} users, described in Section \ref{subsec:content_generation}), we select 50 users who each authored at least $k+m_1$ original tweets (excluding retweets) since 2024\footnote{Note that the dataset extends only up to 2020; therefore, we used the X API to collect more recent tweets.}. This ensures a realistic, unbiased evaluation of model generalization performance on definitively unseen data. The prompt used is identical to Prompt 1 (Appendix~\ref{sec:appendix_B}). 

\noindent \paragraph{Evaluation.}  For this task, we use weighted F1-score as the evaluation metric.


\subsection{Social Media Post Generation}
\label{subsec:content_generation}

In this task, we assess each LLM's capability to create plausible social media posts on behalf of real users. Specifically, the objective is to generate a set of synthetic posts (tweets) that reflect the user's writing style, topical preferences, and social context, drawing on the user's historical posts as guiding examples. 

\paragraph{User Sampling.} We first restrict our pool to users who posted at least 200 \emph{original} tweets (excluding retweets) during the 2018--2020 window. Additionally, we retain only users who have more than 500 followers, more than 100 followees (friends), a non-empty bio with a \text{description} longer than 20 characters, and a \texttt{verified} status. These constraints help ensure each selected user is sufficiently active, has meaningful profile information, and demonstrates personal tweeting behavior. Applying the above filters yields a total of 383 users, each of whom is included in our evaluations for the post generation task. We refer to this set of users as \texttt{VAPOR} users (Verified, Active, Profile-rich, Original). 
Selecting \texttt{VAPOR} users may bias the data toward public-facing accounts; this was intentional for Tasks II–III to ensure high-quality ground truth. In contrast, Task I and the user study (Section~\ref{user_survey} and Appendix~\ref{sec:appendix_E}) include less-active users and casual tweets, ensuring coverage of broader and informal language use.

\paragraph{Prompt Posts.} To generate $n$ new posts for each \texttt{VAPOR} user, we sample $k$ user-authored posts and combine them into a single prompt, along with the user’s description and follower/following statistics (shown as \textit{Other Information} in Figure~\ref{fig:overall}). Each model is instructed to produce a fixed number of posts that resemble the user’s style. We explicitly request short posts not exceeding 280 characters (tweet's character limit), formatted consecutively with minimal additional text.

\paragraph{LLM’s Prompt.} The prompt is included in Appendix~\ref{sec:appendix_B} (Prompt 2). 

\paragraph{Evaluation.} We evaluate the quality of generated posts against two reference sets: \textit{Prompt Posts} and \textit{Non-prompt Posts}. Prompt Posts consist of real user tweets that are included in the input prompt and thus visible to the LLM during generation. In contrast, Non-prompt Posts are a separate set of real tweets authored by the same user but withheld from the prompt, serving as an unseen reference set for evaluation. The purpose is to assess LLMs' capability on ``Example" and ``Evaluation" sets, respectively, similar to what is usually done in machine learning modeling.  As for evaluation measures,  we compute standard natural language generation metrics including BLEU~\cite{papineni2002bleu}, ROUGE~\cite{lin2004rouge} (specifically ROUGE-1 and ROUGE-L) to quantify the lexical overlap with reference tweets. Additionally, we calculate Perplexity~\cite{chen1998evaluation} using GPT-2~\cite{radford2019language} to assess the fluency and naturalness of the generated content. 

\vspace{-12pt}
\begin{figure}[htp]
    \centering
    \includegraphics[width=0.85\linewidth]{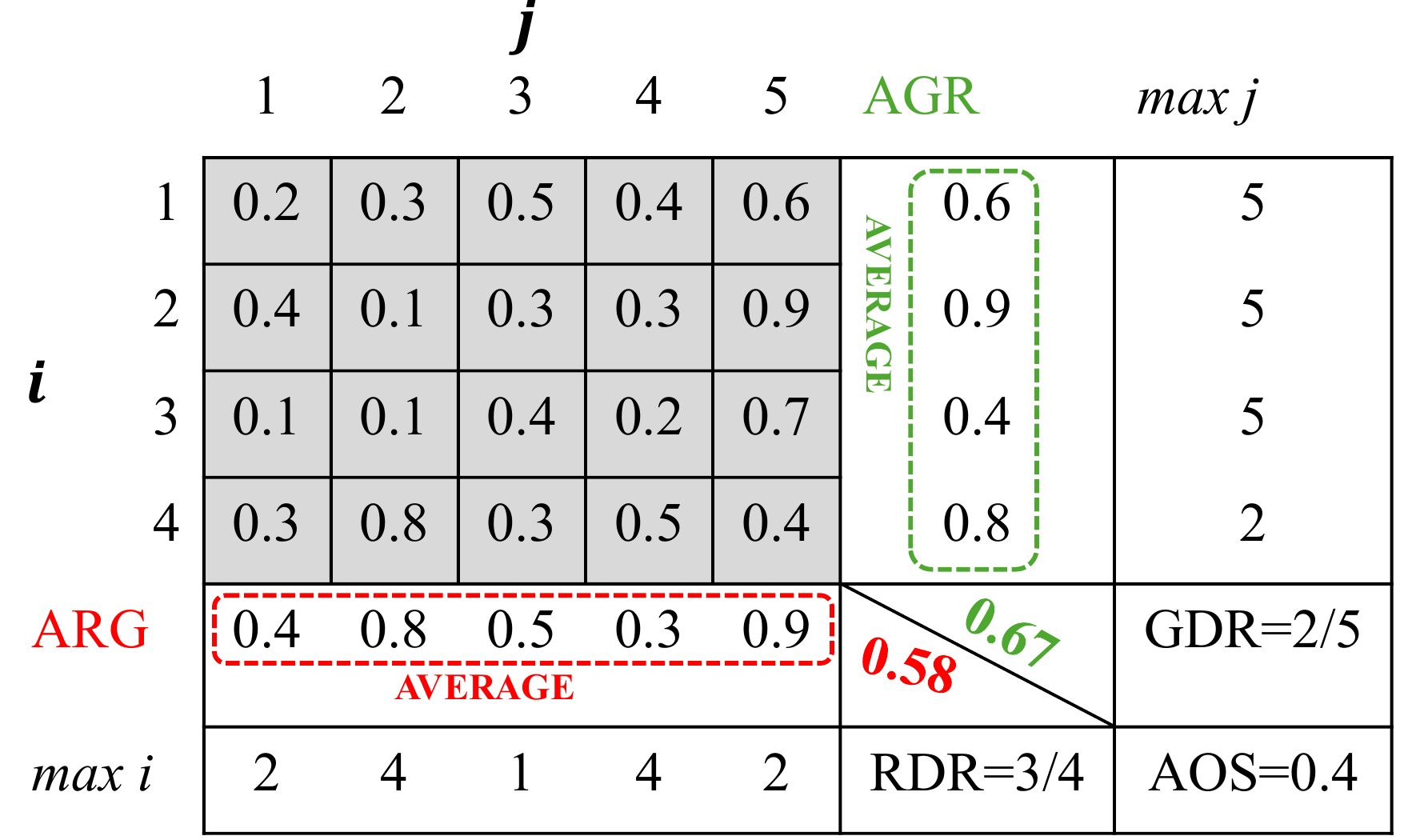}
    \caption{An example of metrics for post generation evaluation}
    \label{fig:post-gen-metrics}
\end{figure}
Moreover, we numerically represent each of the $n$ LLM-generated posts and each of the $k$ posts in our evaluation sets using the SBERT sentence transformer model (\texttt{all-MiniLM-L6-v2})~\cite{reimers2019sentence}. Then, we form an $n \times k$ matrix of $\mathcal{M}$ where entry $\mathcal{M}_{ij}$ stores the cosine similarity between the LLM-generated post $i$ and the real user's post $j$. Using this matrix, we propose the following  metrics (Figure~\ref{fig:post-gen-metrics} demonstrates an example):

\textbf{Average Gen vs. Real (AGR):} For each generated post (row), we retrieve the maximum similarity to any real post. We then take the mean of these maxima across all generated posts. $AGR=\frac{1}{n}\sum_{i=1}^{n}\max_{j}\mathcal{M}_{ij}$.

\textbf{Average Real vs. Gen (ARG):} For each real post (column), we retrieve the maximum similarity to any generated post. We then take the mean of these maxima across all real posts. $ARG=\frac{1}{k}\sum_{j=1}^{k}\max_{i} \mathcal{M}_{ij}$.

 \textbf{Average Overall Similarity (AOS):} The mean of all pairwise cosine similarities between generated and real posts. $AOS=\frac{1}{n\times k}\sum_{i} \sum_{j} \mathcal{M}_{ij}$.
 
\textbf{Gen Dispersion Ratio (GDR):} The fraction of real posts uniquely identified as the most similar match across generated posts (row ratio).

{\small
\[
\text{GDR} = \frac{ \left| unique  \left( \left\{ \underset{j}{\arg\max}~ \mathcal{M}_{ij} \,\middle|\, i = 1, \ldots, n \right \}\right) \right|}{k}
\]
}

\textbf{Real Dispersion Ratio (RDR):} The fraction of generated posts uniquely identified as the most similar match across real posts (column ratio).

{\small
\[
\text{RDR} = \frac{ \left| unique  \left( \left\{ \underset{i}{\arg\max}~ \mathcal{M}_{ij} \,\middle|\, j = 1, \ldots, k \right \}\right) \right|}{n}
\]
}

The last two metrics ($GDR$ and $RDR$) assess the breadth of coverage between generated and real posts: how \textit{broadly} or \textit{narrowly} generated posts cover different real posts, and vice versa.

\subsection{User Study}
\label{user_survey}

Bridging Task~I and Task~II, we conducted a user study to evaluate the perceived authenticity of LLM-generated posts. The study was approved by our university’s IRB (Institutional Review Board), and participants were recruited via an open call. Eligibility required being at least 18 years old, owning a public Twitter/X account, and having at least 50 original tweets. Nineteen users met these criteria and completed the study. Each participant evaluated a personalized set of 20 LLM-generated tweets (five per model from DeepSeek, Gemini, GPT, and Llama) and two authentic tweets randomly sampled from their own timeline, which served as attention checks. Generated tweets were conditioned on the user’s bio, follower/followee counts, and 50 sampled tweets. The exact prompt is provided in Appendix~\ref{sec:appendix_B} (Prompt~2). Participants were instructed as follows:

 \begin{mdframed}[linewidth=1pt,linecolor=black,backgroundcolor=yellow!6,roundcorner=5pt]
 “\textit{The following tweets were generated by AI (LLM) using your publicly available tweets. For each of them, rank how likely it is that you would write it}.”
 \end{mdframed}


Participants rated each tweet using a five-point scale: \textit{Definitely not me}, \textit{Probably not me}, \textit{Unsure}, \textit{Probably me}, and \textit{Definitely me}. If a participant failed to select \textit{Probably me} or \textit{Definitely me} for either of their two real tweets, their responses were deemed unreliable and excluded. This left us \(12\) valid participants. Surveys were administered individually via the Qualtrics platform, with each participant receiving a \$15 Amazon gift card. We also collected basic demographic and Twitter (X) usage information of the participants, summarized in Appendix~\ref{sec:appendix_E}, Table~\ref{tab:user_demographics}.

\subsection{User Attribute Inference}
\label{subsec:user_demographics}

To categorize user attributes in our study, we utilize two formal taxonomies. First, we draw on the \textit{IAB Tech Lab Content Taxonomy v3.1}~\cite{iabtechlab}, a widely adopted standard in digital advertising that provides a hierarchical classification of online content across diverse topics (e.g., news, sports, business, etc.). Specifically, we use the top-level categories of this taxonomy. Second, we use the \textit{2018 Standard Occupational Classification (SOC) System}~\cite{soc}, which is maintained by the U.S.\ Bureau of Labor Statistics to systematically classify occupations in the United States. Using these categories, we annotated and labeled the occupations and interests/hobbies of our \texttt{VAPOR} users (described in Section~\ref{subsec:content_generation}).


Two authors of this paper collaboratively annotated each user’s occupation and interests by closely examining profile descriptions and historical tweets, with disagreements resolved by a third author (a senior researcher). For interests, we initially considered 39 categories from the IAB Tech Lab Content Taxonomy, of which 25 were represented among the selected users. For occupations, we relied on the SOC 2018 hierarchy and annotated at two levels--Level~1 (L$_1$) and Level~2 (L$_2$)--comprising 18 and 38 occupational groups, respectively, to balance granularity and coverage. The full category lists are provided in Appendix~\ref{sec:appendix_F} (see Tables~\ref{tab:interest_categories}--\ref{tab:occupation_l2}), and the Task~III prompt is included in Appendix~\ref{sec:appendix_B} (Prompt~3).

\paragraph{Evaluation.} For each user, an LLM yields a category (for both occupations and interests/hobbies). Then, we evaluate the performance against the ground truth user attributes using accuracy and weighted F1-score metrics. 

\section{Experiments}
\label{sec:experiments}

In this section, we describe our experimental design and present detailed evaluations of the LLMs on three main tasks. The experiments were conducted on a system with an AMD EPYC 7513 CPU, 4 NVIDIA RTX A4000 GPUs, and 1 TB of RAM.

\begin{table*}[htp]
  \centering
  \footnotesize
  \setlength{\tabcolsep}{4pt}
    \caption{Experimental results for social media authorship verification, Task I, (metric: weighted F1-score, multiplied by 100 for clarity) }
  \label{tab:accuracy}
  \resizebox{\textwidth}{!}{%
  \begin{tabular}{l@{\hspace{4pt}}*{16}{c}@{\hspace{6pt}}|@{\hspace{6pt}}c@{\hspace{6pt}}|@{\hspace{6pt}}c}
    \toprule
    & \multicolumn{15}{c}{\textbf{Negative Social Media Post Sampling}} & & & \\[-0.4em]
    \cmidrule(lr){2-16}
    \multirow{3}{*}{Model} &
      \multicolumn{3}{c}{Reciprocal} &
      \multicolumn{3}{c}{Followees-only} &
      \multicolumn{3}{c}{Followers-only} &
      \multicolumn{3}{c}{Random} &
      \multicolumn{3}{c}{Topic‑similar} &
      \multirow{3}{*}{Avg} &
      \multirow{3}{*}{\shortstack[c]{Unseen\\ Data  \\(Knowledge\\Cut-off)}} &
     \multirow{3}{*}{\shortstack[c]{Avg \\Rank}} \\[0.1cm]
    & \multicolumn{15}{c}{\textbf{Positive User Sampling}} & & & \\[-0.4em]
    \cmidrule(lr){2-16}
    & Rnd & Rec & Top & Rnd & Rec & Top & Rnd & Rec & Top &
      Rnd & Rec & Top & Rnd & Rec & Top & & & \\[-0.2em]
    \cmidrule(lr){2-4}\cmidrule(lr){5-7}\cmidrule(lr){8-10}\cmidrule(lr){11-13}\cmidrule(lr){14-16}
    \midrule
    GPT-4 & \textbf{85} & \textbf{83} & \textbf{84} & \textbf{78} & \textbf{86} & \textbf{83} & \textbf{74} & \textbf{87} & \textbf{86} & \textbf{94} & \textbf{93} & \textbf{95} & 76 & \textbf{83} & \textbf{80} & \textbf{84.5} & \textbf{85} & 1.07\\
    Gemini & 75 & 81 & 80 & 59 & 78 & 79 & 69 & 78 & 80 & 84 & 85 & 87 & 57 & 72 & 69 & 75.5 & 79 & 2.87\\
    DeepSeek & 73 & 77 & 77 & 68 & 77 & 76 & 72 & 76 & 75 & 86 & 87 & 87 & 56 & 70 & 70 & 75.1 & 80 & 3.47\\
    RF & 70 & 80 & 82 & 63 & 78 & 75 & 64 & 76 & 77 & 59 & 67 & 72 & \textbf{78} & 65 & 65 & 71.4 & - & 4.33\\
    TF-IDF & 60 & 60 & 60 & 60 & 60 & 60 & 57 & 60 & 60 & 71 & 81 & 81 & 57 & 75 & 75 & 65.1 & - & 5.80\\
    Compression-NCD & 59 & 63 & 62 & 59 & 62 & 62 & 57 & 61 & 61 & 65 & 78 & 77 & 55 & 71 & 71 & 64.2 & - & 6.20\\
    SIAMESE + SBERT & 57 & 60 & 59 & 59 & 60 & 60 & 56 & 60 & 60 & 69 & 74 & 74 & 60 & 34 & 34 & 58.0 & - & 7.67\\
    SIAMESE + GloVe & 19 & 75 & 77 & 40 & 70 & 68 & 65 & 7 & 76 & 60 & 62 & 62 & 33 & 34 & 49 & 53.1 & - & 7.87\\
    Llama & 61 & 74 & 76 & 50 & 70 & 71 & 55 & 72 & 70 & 51 & 51 & 51 & 43 & 47 & 44 & 59.1 & 56 & 7.87\\
    GPT-3.5-Turbo & 58 & 61 & 59 & 57 & 56 & 56 & 55 & 57 & 55 & 62 & 60 & 59 & 57 & 57 & 57 & 57.7 & 71 & 8.47\\
    Bert & 55 & 77 & 4 & 27 & 67 & 41 & 51 & 7 & 73 & 41 & 33 & 36 & 43 & 34 & 33 & 41.5 & - & 10.13\\
    USE & 28 & 43 & 43 & 41 & 44 & 44 & 28 & 43 & 44 & 49 & 60 & 59 & 42 & 53 & 53 & 44.9 & - & 10.47\\
    \bottomrule
  \end{tabular}%
  }
\end{table*}

\subsection{Social Media Authorship Verification}
\label{sec:exp authorship}

For this task, we benchmark five LLMs--GPT-4, GPT-3.5-Turbo, Gemini, Llama, and DeepSeek--alongside BERT~\cite{devlin2019bert} and a Random Forest classifier (training details in Appendix~\ref{sec:appendix_C}). We further include established authorship verification baselines: Universal Sentence Encoder (USE) cosine similarity~\cite{cer2018universal}, TF-IDF cosine similarity~\cite{stamatatos2009survey}, a compression-based impostors method using normalized compression distance (NCD)~\cite{potha2017improved}, and two Siamese models based on GloVe-LSTM~\cite{boenninghoff2019similarity} and SBERT embeddings~\cite{reimers2019sentence} (baseline descriptions in Appendix~\ref{sec:appendix_C}). Performance is measured using weighted F1-score, pooling predictions across users. We evaluate 15 settings formed by the Cartesian product of three positive user sampling schemes--\textit{Random}, \textit{Recent Active}, and \textit{Top Active}--and five negative post sampling strategies--\textit{Random}, \textit{Topic-similar}, \textit{Followers-only}, \textit{Followees-only}, and \textit{Reciprocal} (Section~\ref{subsec:authorship_verification}).  The number of \textit{Few-shot Example Posts} and \textit{Positive}/\textit{Negative Evaluation Posts} is fixed ($k=m_1=m_2=20$), and each setting includes 50 users. Table~\ref{tab:accuracy} reports results, including evaluation on unseen data collected after model knowledge cut-off dates.

As shown in Table~\ref{tab:accuracy}, GPT-4 consistently achieves the strongest performance, with an average F1-score of 0.845 and the top rank (1.07). Gemini and DeepSeek form a strong second tier, with average F1-scores of approximately 0.755 and 0.751, respectively. Traditional baselines (e.g., Random Forest, TF-IDF, and Compression-NCD) achieve intermediate performance, while Siamese models (SBERT and GloVe) show moderate capability and sensitivity to sampling conditions. In contrast, Llama, GPT-3.5-Turbo, BERT, and USE exhibit lower and more variable performance, highlighting the robustness of GPT-4, Gemini, and DeepSeek for authorship verification. On unseen data, GPT-4 again leads (0.85), followed by DeepSeek (0.80) and Gemini (0.79), demonstrating strong generalization beyond the training period. 
\begin{table}[H]
  \centering
  \footnotesize
  \setlength{\tabcolsep}{4pt}
    \caption{Impact of positive user sampling and negative post sampling strategies on model accuracy.}
  \label{tab:sampling_effects}
  \begin{tabular}{lcc}
    \toprule
    \textbf{Model} & \textbf{User Effect (pp)} & \textbf{Tweet Effect (pp)} \\
    \midrule
    DeepSeek      & 4.37 & 20.17 \\
    Gemini        & 6.62 & 18.26 \\
    GPT-3.5-Turbo & 4.53 & 10.62 \\
    GPT-4         & 5.53 & 14.65 \\
    Llama     & 9.47 & 21.13 \\
    \midrule
    Average       & 6.10 & 16.97 \\
    \bottomrule
  \end{tabular}
\end{table}
Table~\ref{tab:sampling_effects} further shows that negative post sampling has a substantially larger impact on accuracy ($\approx$17 percentage points) than positive user sampling ($\approx$6 points). Llama and DeepSeek are most sensitive to negative sampling variations, while GPT-3.5-Turbo shows the lowest variability. Overall, these results underscore GPT-4’s robustness and the critical role of negative post sampling in authorship verification. Appendix~\ref{sec:appendix_C} provides supplementary materials for Task~I, including details on the computation of the User and Tweet Effects, qualitative analysis (Table~\ref{tab:authorship_with_history}), class-wise results, and implementation details such as model access and hyperparameters.

\subsection{Social Media Post Generation}

The evaluated models for this task include GPT-4o, Gemini, DeepSeek, and Llama, along with traditional baselines such as the Markov Chain~\cite{shannon1948mathematical,freitas2015reverse}, BART-large~\cite{lewis2019bart}, and T5-large~\cite{raffel2020exploring}. For BART and T5, we used the `large' pretrained variants. We used $k = 50$, $n = 10$, and $50$ Prompt and $50$ Nonprompt posts (Section~\ref{subsec:content_generation}).

\begin{table*}[t]
\centering
\captionsetup{justification=raggedright,singlelinecheck=false}
\caption{\small Experimental results for social media post generation (Task II)}
\label{tab:content_generation_results}
\begin{minipage}{0.70\textwidth}
\centering
\resizebox{\textwidth}{!}{
\begin{tabular}{llccccccccc}
\toprule
\textbf{Set} & \textbf{Model} &
\textbf{AOS} & \textbf{AGR} &
\textbf{ARG} & \textbf{GDR} &
\textbf{RDR} & \textbf{BLEU} &
\textbf{ROUGE-1} & \textbf{ROUGE-L} \\
\midrule
\multirow{7}{*}{\rotatebox{-90}{Prompt}}
 & GPT-4o        & 0.196 & 0.466 & 0.333 & 0.157 & 0.919 & 0.049 & 0.233 & 0.187  \\
 & Gemini  & 0.199 & 0.442 & 0.319 & 0.134 & 0.900 & 0.041 & 0.209 & 0.178 \\
 & DeepSeek        & 0.214 & 0.493 & 0.363 & 0.163 & 0.926 & 0.065 & 0.264 & 0.226 \\
 & Llama     & 0.188 & 0.489 & 0.341 & 0.156 & 0.908 & 0.135 & 0.292 & 0.253 \\
 & Markov Chain    & 0.260 & 0.750 & 0.424 & 0.117 & 0.695 & 0.957 & 0.670 & 0.668 \\
 & BART-large    & 0.257 & 0.516 & 0.386 & 0.140 & 0.800 & 0.099 & 0.241 & 0.179 \\
 & T5-large    & 0.228 & 0.542 & 0.372 & 0.060 & 0.800 & 0.029 & 0.234 & 0.189 \\
\midrule
\multirow{7}{*}{\rotatebox{-90}{Nonprompt}}
 & GPT-4o        & 0.191 & 0.417 & 0.315 & 0.144 & 0.891 & 0.029 & 0.211 & 0.168  \\
 & Gemini  & 0.195 & 0.424 & 0.312 & 0.133 & 0.893 & 0.037 & 0.201 & 0.171 \\
 & DeepSeek        & 0.209 & 0.434 & 0.341 & 0.151 & 0.885 & 0.042 & 0.236 & 0.204 \\
 & Llama      & 0.183 & 0.408 & 0.317 & 0.146 & 0.875 & 0.035 & 0.220 & 0.181 \\
 & Markov Chain    & 0.245 & 0.504 & 0.374 & 0.112 & 0.658 & 0.123 & 0.345 & 0.320 \\
 & BART-large    & 0.215 & 0.438 & 0.330 & 0.120 & 1.000 & 0.006 & 0.183 & 0.130 \\
 & T5-large     & 0.219 & 0.456 & 0.351 & 0.140 & 0.900 & 0.017 & 0.206 & 0.170 \\
 \bottomrule
\end{tabular}}
\end{minipage}
\hfill
\begin{minipage}{0.29\textwidth}
\centering
\captionsetup{justification=centering,singlelinecheck=true}
\caption{\small Perplexity Scores (Task II)}
\label{tab:perplexity_comparison}
\vspace{0.5pt}
\small
\setlength{\tabcolsep}{1pt}
\begin{tabular}{lc}
\toprule
\textbf{Model} & \textbf{Perplexity} \\
\midrule
GPT-4o & 73.43 \\
Gemini & 129.31 \\
DeepSeek & 141.83 \\
Llama & 76.78 \\
Markov Chain & 263.26 \\
BART-large & 1.123 \\
T5-large & 3.00 \\
\bottomrule
\end{tabular}
\end{minipage}
\end{table*}

Tables~\ref{tab:content_generation_results} and~\ref{tab:perplexity_comparison} summarize model behavior in post generation. The Markov Chain baseline shows strong replication (high AOS and BLEU) but very poor fluency (Perplexity: 263.26). Among LLMs, DeepSeek attains the highest overall semantic similarity in the Prompt setting, indicating closer semantic alignment with real posts, yet its high perplexity suggests reduced fluency. Llama achieves the highest BLEU and ROUGE scores among LLMs, reflecting stronger lexical reuse, but records the lowest overall semantic similarity. GPT-4o offers the most balanced profile, combining strong semantic similarity, broad coverage (high RDR), and the lowest perplexity among LLMs. Gemini emphasizes paraphrasing, yielding moderate semantic similarity with comparatively higher perplexity. Transformer baselines (BART-large, T5-large) achieve extremely low perplexity—indicating highly predictable outputs—but show narrower coverage under Prompt conditioning. Overall, the results reveal trade-offs among semantic similarity, fluency, lexical reuse, and coverage breadth.



\begin{figure}[htbp]
    \centering
    \includegraphics[width=\linewidth]{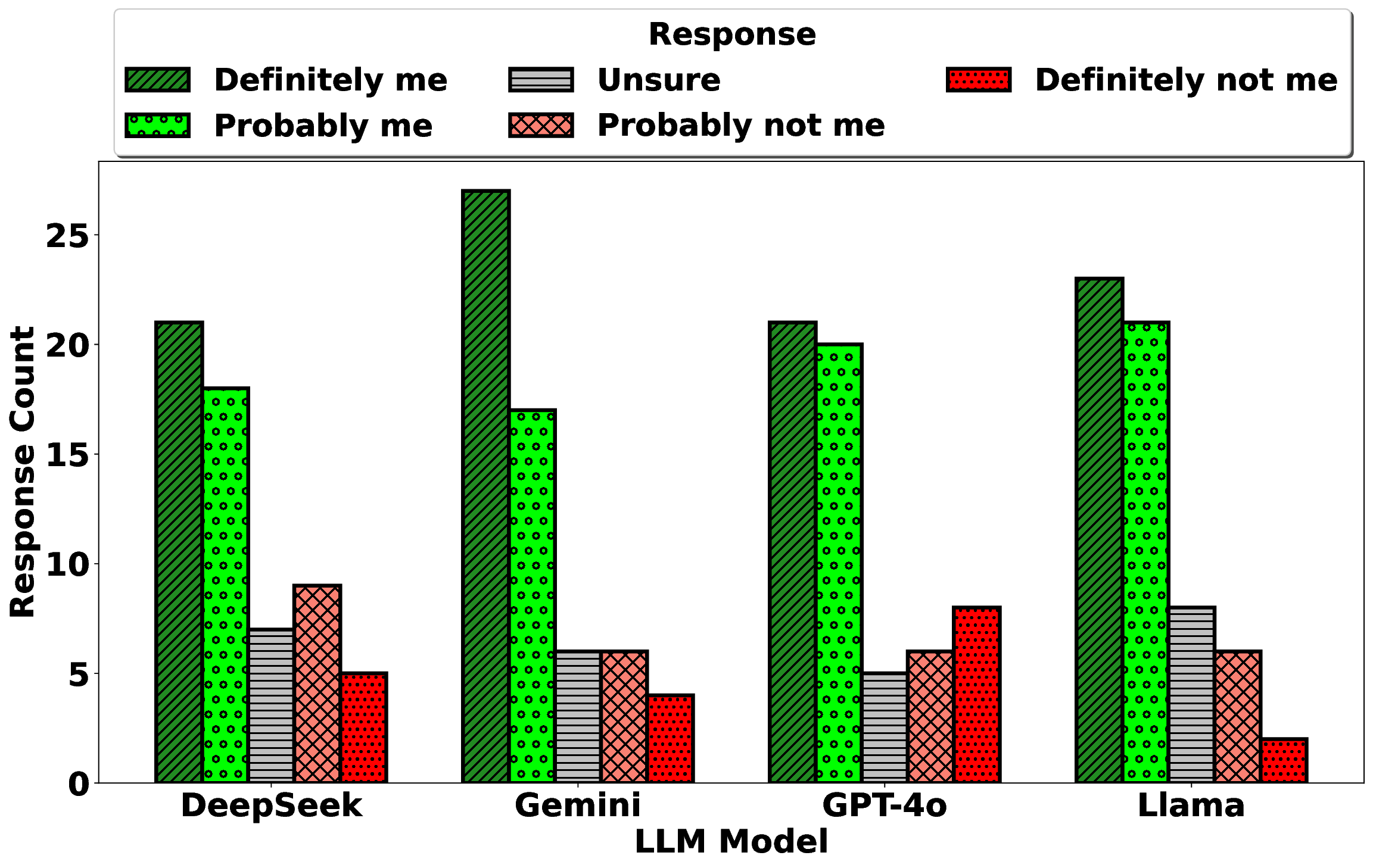}
    \vspace{-20pt}
    \caption{\small User survey responses categorized by generated tweets from different LLMs}
    \label{fig:user_survey_responses}
\end{figure}

\subsection{User Study Results}
\label{ssec:user_survey}





\begin{table*}[!htp]
\centering
\footnotesize
\setlength{\tabcolsep}{5pt}
\caption{Experimental results for user attribute inference (Task III)}
\label{tab:llm_results}
\begin{tabular}{l|cc|cc|cc}
\toprule
\textbf{Model} & \multicolumn{2}{c|}{\textbf{Interests/Hobbies}} & \multicolumn{2}{c|}{\textbf{Occupations (L$_1$)}} & \multicolumn{2}{c}{\textbf{Occupations (L$_2$)}} \\
              & Accuracy & Weighted F1 & Accuracy & Weighted F1 & Accuracy & Weighted F1 \\
\midrule
GPT-4o      & 71.54 & 75.07 & 67.62 & 66.01 & 56.13 & 51.50 \\
Gemini       & \textbf{76.24} & \textbf{76.49} & \textbf{78.32} & \textbf{76.61} & \textbf{61.87} & \textbf{60.83} \\
DeepSeek     & 69.19 & 69.53 & 71.27 & 68.19 & 57.18 & 51.53 \\
Llama        & 30.54 & 36.28 & 35.77 & 43.29 & 2.61 & 1.35 \\
\cite{preoctiuc2015analysis} & 2.60 & 2.33 & 22.08 & 29.27 & 13.04 & 10.32 \\
\cite{lewis2019bart} (Few-shot) & 46.56 & 31.27 & 70.13 & 63.08 & 50.65 & 39.47 \\
\cite{michelson2010discovering} & 15.61 & 8.11 & 27.42 & 3.84 & 12.99 & 5.56 \\
\cite{pennacchiotti2011machine} & 41.56 & 37.75 & 55.84 & 49.58 & 32.47 & 32.12 \\
\bottomrule
\end{tabular}
\end{table*}


To assess perceived authenticity, we conducted the user study described in Section~\ref{user_survey}. Figure~\ref{fig:user_survey_responses} shows the distribution of ratings across five categories. Gemini and Llama receive the highest concentration of positive judgments (\emph{Definitely/Probably me}; 44/60 each), followed by GPT-4o (41/60) and DeepSeek (39/60). GPT-4o also receives the most \emph{Definitely not me} responses, while DeepSeek shows the highest number of \emph{Probably not me} ratings. The neutral option (\emph{Unsure}) remains below 15\% across models, suggesting that participants generally formed clear opinions. Mean authenticity scores further confirm this pattern: Gemini and Llama achieve the highest ratings (3.95/5), with DeepSeek and GPT-4o slightly lower (3.67–3.68). Additional statistical and qualitative analyses are provided in Appendix~\ref{sec:appendix_E} (Tables~\ref{tab:user_survey} and~\ref{tab:heri_llm_coverage}).

\textbf{Note:} All users consented to having both their own tweets and LLM-generated tweets made publicly available.

We further compare human ratings (Appendix~\ref{sec:appendix_E}, Table~\ref{tab:user_survey}) with automatic metrics (Tables~\ref{tab:content_generation_results} and~\ref{tab:perplexity_comparison}) to examine their alignment with perceived authenticity. Llama shows the strongest consistency, combining high author-likeness ratings with strong BLEU and ROUGE scores, suggesting that lexical reuse enhances perceived authenticity. Gemini achieves similarly high human ratings despite lower lexical overlap and higher perplexity, indicating that paraphrastic imitation can also be effective. DeepSeek attains high semantic similarity but lower authenticity ratings, implying that topical alignment alone is insufficient. GPT-4o demonstrates moderate performance across both human and automatic measures, reinforcing that no single metric reliably predicts perceived author-likeness. Overall, the results reveal trade-offs across lexical, stylistic, and semantic dimensions.

\subsection{User Attribute Inference}

We evaluate GPT-4o, Gemini, DeepSeek, and Llama on user attribute inference by predicting the occupations and interests of \texttt{VAPOR} users (Section~\ref{subsec:content_generation}). For each user, 50 sampled tweets are provided as input to the prompt described in Section~\ref{subsec:user_demographics}. We compare LLMs' performance against several baselines, including a Gaussian Process classifier with Word2Vec embeddings~\cite{preoctiuc2015analysis}, few-shot BART~\cite{lewis2019bart}, an entity-based DBpedia classifier~\cite{michelson2010discovering}, and a TF-IDF gradient boosting model~\cite{pennacchiotti2011machine}. Table~\ref{tab:llm_results} shows the results. Performance is measured using accuracy and the weighted F1-score.

Gemini achieves the strongest results across interests and both occupational levels (L$_1$ and L$_2$). GPT-4o and DeepSeek perform comparably at broader levels (L$_1$) but decline at finer granularity (L$_2$), while Llama consistently underperforms. These findings indicate that increasing classification granularity significantly affects LLM accuracy, with Gemini demonstrating the most robust overall performance.

Error patterns are further analyzed via confusion matrices (Table~\ref{tab:task3_confusion}, Appendix~\ref{sec:appendix_F}). Appendix~\ref{sec:appendix_F} also provides implementation details, baseline descriptions, category lists, and qualitative analysis.

\section{Conclusion}
\label{sec:conclusion}

In this paper, we presented a comprehensive evaluation of modern LLMs across three core social media analytics tasks: social media authorship verification, social media post generation, and user attribute inference. GPT-4 achieved the strongest results in authorship verification, particularly under the unseen-data regime. In the post generation, DeepSeek showed strong semantic alignment, while Gemini and Llama received the highest human authenticity ratings. For attribute inference, Gemini performed best, especially at finer-grained hierarchical levels. Overall, our findings highlight the importance of controlled sampling and multifaceted evaluation when assessing LLMs in social media contexts.

Our study has some limitations. Although no other datasets meeting our criteria were identified, future work could extend our methodology to additional datasets and platforms (e.g., Reddit). Moreover, our user study is not large-scale. Future studies should use larger, more diverse samples.




Future work may extend this framework by incorporating multimodal signals (e.g., images, videos, and social graphs) to enrich stylistic and demographic analyses. Modeling reposting and diffusion dynamics could further clarify content virality. More broadly, advancing LLM-based modeling of social network evolution—potentially via link prediction that integrates textual, temporal, and structural signals—would strengthen comprehensive benchmarks in social media analytics.



Finally, we note ethical risks in modeling user identity, including impersonation and privacy-sensitive profiling. This work is intended for benchmarking, not unsafeguarded deployment.

\bibliography{custom}

@article{kumar2024longlamp,
  title={Longlamp: A benchmark for personalized long-form text generation},
  author={Kumar, Ishita and Viswanathan, Snigdha and Yerra, Sushrita and Salemi, Alireza and Rossi, Ryan A and Dernoncourt, Franck and Deilamsalehy, Hanieh and Chen, Xiang and Zhang, Ruiyi and Agarwal, Shubham and others},
  journal={arXiv preprint arXiv:2407.11016},
  year={2024}
}

@inproceedings{salemi2024lamp,
  title={Lamp: When large language models meet personalization},
  author={Salemi, Alireza and Mysore, Sheshera and Bendersky, Michael and Zamani, Hamed},
  booktitle={Proceedings of the 62nd Annual Meeting of the Association for Computational Linguistics (Volume 1: Long Papers)},
  pages={7370--7392},
  year={2024}
}

@article{zhang2024personalization,
  title={Personalization of large language models: A survey},
  author={Zhang, Zhehao and Rossi, Ryan A and Kveton, Branislav and Shao, Yijia and Yang, Diyi and Zamani, Hamed and Dernoncourt, Franck and Barrow, Joe and Yu, Tong and Kim, Sungchul and others},
  journal={arXiv preprint arXiv:2411.00027},
  year={2024}
}

@article{celikyilmaz2020evaluation,
  title={Evaluation of text generation: A survey},
  author={Celikyilmaz, Asli and Clark, Elizabeth and Gao, Jianfeng},
  journal={arXiv preprint arXiv:2006.14799},
  year={2020}
}

@article{perez2023efficiency,
title = {Efficiency of automatic text generators for online review content generation},
journal = {Technological Forecasting and Social Change},
volume = {189},
pages = {122380},
year = {2023},
issn = {0040-1625},
doi = {https://doi.org/10.1016/j.techfore.2023.122380},
url = {https://www.sciencedirect.com/science/article/pii/S0040162523000653},
author = {A. Perez-Castro and M.R. Martínez-Torres and S.L. Toral}
}

@article{singh2018impact,
  title={Impact of social media on e-commerce},
  author={Singh, Manohar and Singh, Gobindbir},
  journal={International Journal of Engineering \& Technology},
  volume={7},
  number={2.30},
  pages={21--26},
  year={2018}
}

@article{schillinger2020infodemics,
  title={From “infodemics” to health promotion: a novel framework for the role of social media in public health},
  author={Schillinger, Dean and Chittamuru, Deepti and Ram{\'\i}rez, A Susana},
  journal={American journal of public health},
  volume={110},
  number={9},
  pages={1393--1396},
  year={2020},
  publisher={American Public Health Association}
}

@article{saroj2020use,
  title={Use of social media in crisis management: A survey},
  author={Saroj, Anita and Pal, Sukomal},
  journal={International Journal of Disaster Risk Reduction},
  volume={48},
  pages={101584},
  year={2020},
  publisher={Elsevier}
}

@inproceedings{hu2024bad,
  title={Bad actor, good advisor: Exploring the role of large language models in fake news detection},
  author={Hu, Beizhe and Sheng, Qiang and Cao, Juan and Shi, Yuhui and Li, Yang and Wang, Danding and Qi, Peng},
  booktitle={Proceedings of the AAAI Conference on Artificial Intelligence},
  volume={38},
  number={20},
  pages={22105--22113},
  year={2024}
}

@article{mu2024large,
  title={Large language models offer an alternative to the traditional approach of topic modelling},
  author={Mu, Yida and Dong, Chun and Bontcheva, Kalina and Song, Xingyi},
  journal={arXiv preprint arXiv:2403.16248},
  year={2024}
}

@article{zhang2023sentiment,
  title={Sentiment analysis in the era of large language models: A reality check},
  author={Zhang, Wenxuan and Deng, Yue and Liu, Bing and Pan, Sinno Jialin and Bing, Lidong},
  journal={arXiv preprint arXiv:2305.15005},
  year={2023}
}

@inproceedings{boenninghoff2019similarity,
  title={Similarity learning for authorship verification in social media},
  author={Boenninghoff, Benedikt and Nickel, Robert M and Zeiler, Steffen and Kolossa, Dorothea},
  booktitle={ICASSP 2019-2019 IEEE international conference on acoustics, speech and signal processing (ICASSP)},
  pages={2457--2461},
  year={2019},
  organization={IEEE}
}

@article{tyo2022state,
  title={On the state of the art in authorship attribution and authorship verification},
  author={Tyo, Jacob and Dhingra, Bhuwan and Lipton, Zachary C},
  journal={arXiv preprint arXiv:2209.06869},
  year={2022}
}

@article{zhang2024benchmarking,
  title={Benchmarking large language models for news summarization},
  author={Zhang, Tianyi and Ladhak, Faisal and Durmus, Esin and Liang, Percy and McKeown, Kathleen and Hashimoto, Tatsunori B},
  journal={Transactions of the Association for Computational Linguistics},
  volume={12},
  pages={39--57},
  year={2024},
  publisher={MIT Press One Broadway, 12th Floor, Cambridge, Massachusetts 02142, USA~…}
}

@inproceedings{huang2024can,
  title={Can Large Language Models Identify Authorship?},
  author={Huang, Baixiang and Chen, Canyu and Shu, Kai},
  booktitle={Findings of the Association for Computational Linguistics: EMNLP 2024},
  year={2024}
}

@article{huang2025authorship,
  title={Authorship attribution in the era of llms: Problems, methodologies, and challenges},
  author={Huang, Baixiang and Chen, Canyu and Shu, Kai},
  journal={ACM SIGKDD Explorations Newsletter},
  volume={26},
  number={2},
  pages={21--43},
  year={2025},
  publisher={ACM New York, NY, USA}
}

@inproceedings{yu2024repalm,
  title={RePALM: Popular Quote Tweet Generation via Auto-Response Augmentation},
  author={Yu, Erxin and Li, Jing and Xu, Chunpu},
  booktitle={Findings of the Association for Computational Linguistics ACL 2024},
  pages={9566--9579},
  year={2024}
}

@inproceedings{wen2023towards,
  title={Towards open-domain Twitter user profile inference},
  author={Wen, Haoyang and Xiao, Zhenxin and Hovy, Eduard and Hauptmann, Alexander G},
  booktitle={Findings of the Association for Computational Linguistics: ACL 2023},
  pages={3172--3188},
  year={2023}
}

@article{zhao2025amplifying,
  title={Amplifying Your Social Media Presence: Personalized Influential Content Generation with LLMs},
  author={Zhao, Yuying and Wang, Yu and Cheng, Xueqi and Tumlin, Anne Marie and Liu, Yunchao and Xia, Damin and Jiang, Meng and Derr, Tyler},
  journal={arXiv preprint arXiv:2505.01698},
  year={2025}
}

@inproceedings{pillai2025engagement,
  title={Engagement-driven Persona Prompting for Rewriting News Tweets},
  author={Pillai, Reshmi Gopalakrishna and Fokkens, Antske and van Atteveldt, Wouter},
  booktitle={Proceedings of the 31st International Conference on Computational Linguistics},
  pages={8612--8622},
  year={2025}
}

@article{gambini2024evaluating,
  title={Evaluating large language models for user stance detection on X (Twitter)},
  author={Gambini, Margherita and Senette, Caterina and Fagni, Tiziano and Tesconi, Maurizio},
  journal={Machine Learning},
  volume={113},
  number={10},
  pages={7243--7266},
  year={2024},
  publisher={Springer}
}

@inproceedings{kheiri2023analysis,
  title={An analysis of the dynamics of ties on twitter},
  author={Kheiri, Kiana and Khan, Muhammad Fawad Akbar and Derr, Tyler and Karimi, Hamid},
  booktitle={2023 IEEE International Conference on Big Data (BigData)},
  pages={5809--5817},
  year={2023},
  organization={IEEE}
}

@misc{iabtechlab,
  title        = {Content Taxonomy: v3.1},
  author       = {{IAB Tech Lab}},
  howpublished = {\url{https://iabtechlab.com/standards/content-taxonomy/}},
  year         = {2023},
  note         = {Accessed: 2025-05-07}
}

@misc{soc,
  title        = {Standard Occupational Classification (SOC) System},
  author       = {{U.S. Bureau of Labor Statistics}},
  howpublished = {\url{https://www.bls.gov/soc/}},
  year         = {2018},
  note         = {Accessed: 2025-05-07}
}

@article{reimers2019sentence,
  title={Sentence-bert: Sentence embeddings using siamese bert-networks},
  author={Reimers, Nils and Gurevych, Iryna},
  journal={arXiv preprint arXiv:1908.10084},
  year={2019}
}

@article{radford2019language,
  title={Language models are unsupervised multitask learners},
  author={Radford, Alec and Wu, Jeffrey and Child, Rewon and Luan, David and Amodei, Dario and Sutskever, Ilya and others},
  journal={OpenAI blog},
  volume={1},
  number={8},
  pages={9},
  year={2019}
}

@article{du2023automatic,
  title={Automatic text generation using deep learning: providing large-scale support for online learning communities},
  author={Du, Hanxiang and Xing, Wanli and Pei, Bo},
  journal={Interactive Learning Environments},
  volume={31},
  number={8},
  pages={5021--5036},
  year={2023},
  publisher={Taylor \& Francis}
}

@article{Hong2021,
  title = {Enhancing Personalized Ads Using Interest Category Classification of SNS Users Based on Deep Neural Networks},
  author = {Hong, Taekeun and Choi, Jin-A and Lim, Kiho and Kim, Pankoo},
  journal = {Sensors},
  volume = {21},
  number = {1},
  pages = {199},
  year = {2021},
  publisher = {MDPI}
}

@article{Liu2024,
  title={Occupation Prediction with Multimodal Learning from Tweet Messages and Google Street View Images},
  author={Liu, Xinyi and Peng, Bo and Wu, Meiliu and Wang, Mingshu and Cai, Heng and Huang, Qunying},
  journal={AGILE: GIScience Series},
  volume={5},
  pages={36},
  year={2024},
  publisher={Copernicus Publications G{\"o}ttingen, Germany}
}

@article{hu2024instructav,
  title={InstructAV: Instruction Fine-tuning Large Language Models for Authorship Verification},
  author={Hu, Yujia and Hu, Zhiqiang and Seah, Chun-Wei and Lee, Roy Ka-Wei},
  journal={arXiv preprint arXiv:2407.12882},
  year={2024}
}

@article{huertas2024understanding,
  title={Understanding writing style in social media with a supervised contrastively pre-trained transformer},
  author={Huertas-Tato, Javier and Mart{\'\i}n, Alejandro and Camacho, David},
  journal={Knowledge-Based Systems},
  volume={296},
  pages={111867},
  year={2024},
  publisher={Elsevier}
}

@article{alsanoosy2024authorship,
  title={Authorship Attribution for English Short Texts},
  author={Alsanoosy, Tawfeeq and Shalbi, Bodor and Noor, Ayman},
  journal={Engineering, Technology \& Applied Science Research},
  volume={14},
  number={5},
  pages={16419--16426},
  year={2024}
}

@article{qiu2025can,
  title={Can LLMs Simulate Social Media Engagement? A Study on Action-Guided Response Generation},
  author={Qiu, Zhongyi and Lyu, Hanjia and Xiong, Wei and Luo, Jiebo},
  journal={arXiv preprint arXiv:2502.12073},
  year={2025}
}

@inproceedings{zhang2023twhin,
  title={Twhin-bert: A socially-enriched pre-trained language model for multilingual tweet representations at twitter},
  author={Zhang, Xinyang and Malkov, Yury and Florez, Omar and Park, Serim and McWilliams, Brian and Han, Jiawei and El-Kishky, Ahmed},
  booktitle={Proceedings of the 29th ACM SIGKDD conference on knowledge discovery and data mining},
  pages={5597--5607},
  year={2023}
}

@inproceedings{hu2007demographic,
  title={Demographic prediction based on user's browsing behavior},
  author={Hu, Jian and Zeng, Hua-Jun and Li, Hua and Niu, Cheng and Chen, Zheng},
  booktitle={Proceedings of the 16th international conference on World Wide Web},
  pages={151--160},
  year={2007}
}

@article{de2010predicting,
  title={Predicting website audience demographics forweb advertising targeting using multi-website clickstream data},
  author={De Bock, KoenW and Van den Poel, Dirk},
  journal={Fundamenta Informaticae},
  volume={98},
  number={1},
  pages={49--70},
  year={2010},
  publisher={SAGE Publications Sage UK: London, England}
}

@article{goel2014predicting,
  title={Predicting individual behavior with social networks},
  author={Goel, Sharad and Goldstein, Daniel G},
  journal={Marketing Science},
  volume={33},
  number={1},
  pages={82--93},
  year={2014},
  publisher={INFORMS}
}

@article{injadat2016data,
  title={Data mining techniques in social media: A survey},
  author={Injadat, MohammadNoor and Salo, Fadi and Nassif, Ali Bou},
  journal={Neurocomputing},
  volume={214},
  pages={654--670},
  year={2016},
  publisher={Elsevier}
}

@inproceedings{papineni2002bleu,
  title={Bleu: a method for automatic evaluation of machine translation},
  author={Papineni, Kishore and Roukos, Salim and Ward, Todd and Zhu, Wei-Jing},
  booktitle={Proceedings of the 40th annual meeting of the Association for Computational Linguistics},
  pages={311--318},
  year={2002}
}

@inproceedings{lin2004rouge,
  title={Rouge: A package for automatic evaluation of summaries},
  author={Lin, Chin-Yew},
  booktitle={Text summarization branches out},
  pages={74--81},
  year={2004}
}

@article{chen1998evaluation,
  title={Evaluation metrics for language models},
  author={Chen, Stanley F and Beeferman, Douglas and Rosenfeld, Roni},
  year={1998},
  publisher={Carnegie Mellon University}
}

@inproceedings{cer2018universal,
  title={Universal sentence encoder for English},
  author={Cer, Daniel and Yang, Yinfei and Kong, Sheng-yi and Hua, Nan and Limtiaco, Nicole and John, Rhomni St and Constant, Noah and Guajardo-Cespedes, Mario and Yuan, Steve and Tar, Chris and others},
  booktitle={Proceedings of the 2018 conference on empirical methods in natural language processing: system demonstrations},
  pages={169--174},
  year={2018}
}

@article{stamatatos2009survey,
  title={A survey of modern authorship attribution methods},
  author={Stamatatos, Efstathios},
  journal={Journal of the American Society for information Science and Technology},
  volume={60},
  number={3},
  pages={538--556},
  year={2009},
  publisher={Wiley Online Library}
}

@inproceedings{potha2017improved,
  title={An improved impostors method for authorship verification},
  author={Potha, Nektaria and Stamatatos, Efstathios},
  booktitle={International conference of the cross-language evaluation forum for European languages},
  pages={138--144},
  year={2017},
  organization={Springer}
}

@article{shannon1948mathematical,
  title={A mathematical theory of communication},
  author={Shannon, Claude E},
  journal={The Bell system technical journal},
  volume={27},
  number={3},
  pages={379--423},
  year={1948},
  publisher={Nokia Bell Labs}
}

@inproceedings{devlin2019bert,
  title={Bert: Pre-training of deep bidirectional transformers for language understanding},
  author={Devlin, Jacob and Chang, Ming-Wei and Lee, Kenton and Toutanova, Kristina},
  booktitle={Proceedings of the 2019 conference of the North American chapter of the association for computational linguistics: human language technologies, volume 1 (long and short papers)},
  pages={4171--4186},
  year={2019}
}

@inproceedings{freitas2015reverse,
  title={Reverse engineering socialbot infiltration strategies in twitter},
  author={Freitas, Carlos and Benevenuto, Fabricio and Ghosh, Saptarshi and Veloso, Adriano},
  booktitle={Proceedings of the 2015 IEEE/ACM International Conference on Advances in Social Networks Analysis and Mining 2015},
  pages={25--32},
  year={2015}
}

@article{lewis2019bart,
  title={BART: Denoising sequence-to-sequence pre-training for natural language generation, translation, and comprehension},
  author={Lewis, Mike and Liu, Yinhan and Goyal, Naman and Ghazvininejad, Marjan and Mohamed, Abdelrahman and Levy, Omer and Stoyanov, Ves and Zettlemoyer, Luke},
  journal={arXiv preprint arXiv:1910.13461},
  year={2019}
}

@article{raffel2020exploring,
  title={Exploring the limits of transfer learning with a unified text-to-text transformer},
  author={Raffel, Colin and Shazeer, Noam and Roberts, Adam and Lee, Katherine and Narang, Sharan and Matena, Michael and Zhou, Yanqi and Li, Wei and Liu, Peter J},
  journal={Journal of machine learning research},
  volume={21},
  number={140},
  pages={1--67},
  year={2020}
}

@inproceedings{preoctiuc2015analysis,
  title={An analysis of the user occupational class through Twitter content},
  author={Preo{\c{t}}iuc-Pietro, Daniel and Lampos, Vasileios and Aletras, Nikolaos},
  booktitle={Proceedings of the 53rd Annual Meeting of the Association for Computational Linguistics and the 7th International Joint Conference on Natural Language Processing (Volume 1: Long Papers)},
  pages={1754--1764},
  year={2015}
}

@inproceedings{michelson2010discovering,
  title={Discovering users' topics of interest on twitter: a first look},
  author={Michelson, Matthew and Macskassy, Sofus A},
  booktitle={Proceedings of the fourth workshop on Analytics for noisy unstructured text data},
  pages={73--80},
  year={2010}
}

@inproceedings{pennacchiotti2011machine,
  title={A machine learning approach to twitter user classification},
  author={Pennacchiotti, Marco and Popescu, Ana-Maria},
  booktitle={Proceedings of the international AAAI conference on web and social media},
  volume={5},
  number={1},
  pages={281--288},
  year={2011}
}

@incollection{shulman2022self,
  title={Self-presentation: Impression management in the digital age},
  author={Shulman, David},
  booktitle={The Routledge international handbook of Goffman studies},
  pages={26--37},
  year={2022},
  publisher={Routledge}
}

@article{gunduz2017effect,
  title={The effect of social media on identity construction},
  author={G{\"u}nd{\"u}z, U{\u{g}}ur},
  journal={Mediterranean journal of social sciences},
  volume={8},
  number={5},
  year={2017}
}

@article{yadav2017social,
  title={Social media writing style fingerprint},
  author={Yadav, Himank and Li, Juliang},
  journal={arXiv preprint arXiv:1712.04762},
  year={2017}
}

@article{abbasi2008writeprints,
  title={Writeprints: A stylometric approach to identity-level identification and similarity detection in cyberspace},
  author={Abbasi, Ahmed and Chen, Hsinchun},
  journal={ACM Transactions on Information Systems (TOIS)},
  volume={26},
  number={2},
  pages={1--29},
  year={2008},
  publisher={ACM New York, NY, USA}
}

@inproceedings{tigunova2020reddust,
  title={RedDust: a large reusable dataset of Reddit user traits},
  author={Tigunova, Anna and Mirza, Paramita and Yates, Andrew and Weikum, Gerhard},
  booktitle={Proceedings of the Twelfth Language Resources and Evaluation Conference},
  pages={6118--6126},
  year={2020}
}

@article{chawla2002smote,
  title={SMOTE: synthetic minority over-sampling technique},
  author={Chawla, Nitesh V and Bowyer, Kevin W and Hall, Lawrence O and Kegelmeyer, W Philip},
  journal={Journal of artificial intelligence research},
  volume={16},
  pages={321--357},
  year={2002}
}

\newpage
\appendix
\section*{Appendices}
\section{Dataset}
\label{sec:appendix_A}

We performed our evaluations using the Twitter dataset introduced by~\cite{kheiri2023analysis}, comprising data from over 120,000 users tracked across a 15-week period. This dataset is particularly suitable due to its large scale, diversity, and comprehensive metadata and content coverage. It includes weekly snapshots of user networks, capturing the dynamic formation and dissolution of follower connections. Additionally, it provides detailed user-generated content (tweets), interactions (such as mentions), and extensive user profile attributes, including verification status and user activity metrics. The dataset captures an average of 1,175,846 new tweets weekly and approximately 2.9 million total social ties, making it ideal for our comprehensive analysis. We have explicitly obtained permission from the original authors to utilize this dataset in our study. The dataset statistics are summarized in Table~\ref{tab:dataset_stats}.

\begin{table}[htbp]
\centering
\caption{Twitter dataset statistics acquired from~\cite{kheiri2023analysis}}
\label{tab:dataset_stats}
\resizebox{0.38\textwidth}{!}{%
\small
\begin{tabular}{lr}
\toprule
\textbf{Network Property} & \textbf{Value} \\
\midrule
Total users & 123,829 \\
Total ties & 2,922,732 \\
\# Verified accounts & 3,829 \\
Avg. weekly new followers & 10,855 \\
Avg. weekly new unfollowers & 465 \\
Avg. weekly new Tweets & 1,175,846 \\
Percentage verified users & 1.687\\
Avg. followees count per user & 205 \\
Avg. followers count per user & 150 \\
Diameter (longest shortest path) & 8 \\
Avg. new tweets (w/mentions) & 2,021 \\
\bottomrule
\end{tabular}%
}
\end{table}

\section{LLMs' Prompts and Expenses}
\label{sec:appendix_B}

\promptbox{1}{%
  \raggedright\small\ttfamily
  User Description:\par
  \{\textlangle description\textrangle\}\par
  \medskip
  Previous Tweets:\par
  \{\textlangle previous\_tweets\textrangle\}\par
  \medskip
  Tweet to Classify:\par
  \{\textlangle tweet\_to\_classify\textrangle\}\par
  \medskip
  Task:\par
  Based on the description and the previous tweets of the user, output `0' if the
  above tweet was \emph{not} generated by the user or `1' if it \emph{was}.  Even
  if the tweet is a URL, output only `0' or `1' as your response.
}{Social media authorship verification prompt (Task I)}

Here, \{description\} and \{previous\_tweets\} correspond to the user’s bio and a few-shot list of the user’s previously authored tweets (excluding retweets). Meanwhile, \{tweet\_to\_classify\} is drawn from either the target user’s own future tweets (Positive Evaluation Posts) or from other users’ tweets (Negative Evaluation Posts). The LLM outputs either \texttt{0} or \texttt{1}, indicating whether it believes the specified tweet belongs to the same user.\\

\promptbox{2}{%
  \raggedright\small\ttfamily
  We have collected $K$ tweets from one user on Twitter, included below:\par
  \medskip
  <tweets go here>\par
  \medskip
  Below, you can also find the user-written bio (description) as well as the number
  of followings and followers of the user:\par
  \medskip
  User Bio: \textlangle user\_description\textrangle\par
  Number of Followings: \textlangle following\_count\textrangle\par
  Number of Followers: \textlangle followers\_count\textrangle\par
  \medskip
  Now, generate exactly \textlangle NUM\_GENERATED\_TWEETS\textrangle\ tweets that
  the user could have posted on their Twitter account. Each tweet should not exceed
  280 characters and must be formatted as:\par
  1.~<tweet\_text>\par
  2.~<tweet\_text>\par
  …\par
  Ensure all \textlangle NUM\_GENERATED\_TWEETS\textrangle\ tweets are included,
  without extra spacing or missing tweets.
}{Social media Post generation prompt (Task II)}

 

For inferring the users' interests and hobbies, we provide the LLMs with the following structured prompt:

\promptbox{3}{%
  \raggedright\small\ttfamily
  Your task is to infer the primary \textlangle ATTRIBUTE\textrangle\ of a
  Twitter/X user based on their tweets.\par
  \medskip
  You will be provided with exactly \textlangle\#SAMPLED\_TWEETS\textrangle\
  randomly selected tweets posted by this user.\par
  \medskip
  Assign the user to exactly one of the following \textlangle ATTRIBUTE\textrangle\
  categories, based on the content of their tweets:\par
  \textlangle CATEGORY\_LIST\textrangle\par
  \medskip
  Here are the user's sampled tweets:\par
  \textlangle SAMPLED\_TWEETS\textrangle\par
  \medskip
  Respond strictly and exclusively in the following JSON format:\par
  \{\,\newline
  \quad "\textlangle JSON\_KEY\textrangle": "\textlangle one category from the
  above list\textrangle"\newline
  \}\par
  \medskip
  Do \emph{not} include any other text, explanation, or formatting.
}{User attribute inference prompt (Task III)}

We instantiate Prompt 3 for (i) interests/hobbies, (ii) SOC 2018 Level 1 occupations, and (iii) SOC 2018 Level 2 occupations by substituting the placeholders in Table~\ref{tab:demographic_placeholders}.

\begin{table*}[htp]
\centering
\caption{Placeholders for interest and occupation prompt.}
\label{tab:demographic_placeholders}
\small
\setlength{\tabcolsep}{4pt}  
\begin{tabularx}{\textwidth}{@{}l l X l@{}}
\toprule
\textbf{Task} & \textbf{\texttt{ATTRIBUTE}} & \textbf{\texttt{CATEGORY\_LIST}} & \textbf{\texttt{JSON\_KEY}} \\ 
\midrule
Interests / hobbies & \texttt{interest category} & IAB categories present in our data & \texttt{interest\_category} \\[2pt]
Occupation L1 & \texttt{occupational category (Level--1)} & 18 SOC major groups & \texttt{occupation\_category} \\[2pt]
Occupation L2 & \texttt{occupational category (Level--2)} & 38 SOC minor groups & \texttt{occupation\_category} \\
\bottomrule
\end{tabularx}
\end{table*}

\subsection*{LLMs' Costs}
Table~\ref{tab:model_costs} provides estimated expenses incurred from using LLMs in our experiments. The total estimated cost for all model experiments across all three tasks is \$220.

\begin{table}[!t]
\centering
\caption{\small Estimated cost of using LLMs in our experiments.}
\label{tab:model_costs}
\small
\renewcommand{\arraystretch}{1.3}
\begin{tabular}{lc}
\toprule
\textbf{Model} & \textbf{Cost (USD)} \\ 
\midrule
GPT-4            & \$150     \\
GPT-4o           & \$20      \\
GPT-3.5-Turbo    & \$20      \\
Gemini           & \$20      \\
DeepSeek         & \$10      \\
Llama            & Free    \\
\midrule
\textbf{Total}   & \textbf{\$220} \\
\bottomrule
\end{tabular}
\end{table}

\section{Social Media Authorship Verification}
\label{sec:appendix_C}

\subsection*{Model Access and Hyperparameters} We accessed LLMs through their respective APIs—OpenAI's \texttt{ChatCompletion} endpoint for GPT-4o, Google's \texttt{GenerativeModel} API for Gemini, and the \texttt{ollama.chat} interface for Llama models. For consistency across models, we fix the temperature parameter to \texttt{0} in all LLM calls, ensuring deterministic outputs in the binary classification setting.

\subsubsection*{Qualitative Analysis for Task I (Case Studies)}  
As shown in Table~\ref{tab:authorship_with_history}, the user's previous tweets reveal a clear and consistent pattern of sharing educational YouTube content explicitly linked to the brand ``Letstute." Gemini, GPT-4o, and DeepSeek effectively recognize this pattern, correctly identifying tweets associated with ``Letstute" educational videos as authentic. GPT-3.5 struggles because it relies heavily on exact wording matches and misses broader topical connections. Llama mistakenly accepts short conversational tweets unrelated to the user's known educational content, indicating difficulty in grasping the user's overall topical theme.

\begin{table*}[!t]
\centering
\caption{ Authorship verification example with model predictions. (GT: Ground Truth, \cmark: correct, \xmark: incorrect)}
\label{tab:authorship_with_history}

\renewcommand{\arraystretch}{1.2}

\begin{minipage}{\textwidth}
\textbf{Previous tweets from the user:}
\begin{itemize}
  \item I liked a @YouTube video \url{https://youtube.com/watch?v=IWX2MobGsnE} Amazing Trick To Understand Arithmetic Progression Formula.
  \item How to present an Answer | Smart Answer Presentation | Letstute: \url{https://youtube.com/watch?v=UH5fspzdRDc}
  \item GST (Goods \& Service Tax) | Problem Solving | LetsTute: \url{https://youtube.com/watch?v=g\_HIDyeAmDI}
\end{itemize}
\end{minipage}

\vspace{2pt}

\begin{tabularx}{\textwidth}{l >{\raggedright\arraybackslash}X c c c c c c}
\toprule
 & \textbf{Tweet Example} & \textbf{GT} & \textbf{Gem.} & \textbf{4o} & \textbf{3.5} & \textbf{Lla.} & \textbf{DS}\\
\midrule

T1 & Revise your syllabus CBSE class 10th as per NCERT solution \url{https://youtube.com/watch?a\&v=aPGgIvHGuQM}
& 1 & \cmark & \cmark & \xmark & \cmark & \cmark \\

T2 & 3 tips to speak English fluently \url{https://youtube.com/watch?a\&v=bQgQSF5HP\_c}
& 1 & \cmark & \cmark & \xmark & \cmark & \cmark \\

T3 & @tvserieshub She can not!
& 0 & \cmark & \cmark & \cmark & \xmark & \cmark \\

T4 & @evanwolf No, I was still doing the actual writing on the computer.
& 0 & \cmark & \cmark & \cmark & \xmark & \cmark \\

\bottomrule
\end{tabularx}

\end{table*}

\subsection*{Random Forest Baseline Implementation} For the authorship verification task, we trained a Random Forest (RF) classifier separately for each dataset configuration, using the same user and tweet splits as in the LLM experiments. Specifically, for each user, we constructed a balanced, pairwise training set by (i) pairing each of the user's $k=20$ Few-shot Example Posts with every other Few-shot Example Post from the same user, yielding $k\times(k-1)=380$ positive pairs per user (labeled as 1), and (ii) pairing each of these $k$ Few-shot Example Posts with $m_2=20$ Negative Evaluation Posts sampled from other users within the same timeframe, resulting in an additional $k\times m_2=400$ negative pairs per user (labeled as 0). For the test set, we paired each user's $k=20$ Few-shot Example Post with their $m_1=20$ Positive Evaluation Posts (labeled as 1) and another set of $m_2=20$ Negative Evaluation Posts from other users (labeled as 0), ensuring no overlap and preventing train–test leakage. Each tweet was embedded once using SBERT \texttt{all-MiniLM-L6-v2}, and embeddings of each tweet pair were concatenated into a single 768-dimensional feature vector. For dataset configurations involving graph-based sampling (\textit{Followers-only}, \textit{Followees-only}, \textit{Reciprocal}), potential class imbalance in the training set was addressed using the Synthetic Minority Over-Sampling Technique (SMOTE)~\cite{chawla2002smote}. Each  RF model was tuned independently via 3-fold grid search, optimizing the hyperparameters detailed in Table~\ref{tab:rf_grid}.

At test time, given a candidate tweet $f$, we formed $k=20$ pairs by individually pairing it with each prior tweet: $(t_1, f), (t_2, f), \dots, (t_k, f)$. The RF predicted labels independently for each pair, resulting in a set of pairwise predictions $P = \{p_1, p_2, \dots, p_k\}$, where each $p_i \in \{0,1\}$. The final classification for the candidate tweet was determined by majority voting:
\[
\hat{y}_f = 
\begin{cases}
1 & \text{if}\quad\sum_{i=1}^{k} p_i > \frac{k}{2} \\[6pt]
0 & \text{otherwise}
\end{cases}
\]

The predicted label $\hat{y}_f$ indicates whether the candidate tweet $f$ is considered authored by the target user ($\hat{y}_f = 1$) or by a different user ($\hat{y}_f = 0$). Similar to the LLM experiments, we evaluated this model's performance using accuracy and weighted F1-scores, ensuring direct comparability with the results obtained from the LLMs.

\begin{table}[ht]
\centering
\small
\caption{Search space for the 3-fold RF hyper-parameter tuning}
\begin{tabular}{@{}ll@{}}
\toprule
\textbf{Hyperparameter} & \textbf{Grid values} \\
\midrule
$n_{\mathrm{estimators}}$        & \{100, 200\} \\
$\text{max\_depth}$              & \{None, 10, 30\} \\
$\text{min\_samples\_split}$     & \{2, 5\} \\
$\text{min\_samples\_leaf}$      & \{1, 3\} \\
\bottomrule
\end{tabular}

\label{tab:rf_grid}
\end{table}

\subsection*{Descriptions of Additional Baseline Models} To complement our LLM-based evaluation, we implemented a diverse set of established authorship verification baselines from the literature. Below is a summary of each:
\begin{itemize}

 \item\textbf{Universal Sentence Encoder (USE) + Cosine Similarity}:  
This semantic similarity approach uses Google's pretrained Universal Sentence Encoder to embed tweets. An author's representation is formed by averaging the embeddings of their prior tweets. Cosine similarity is then computed between this profile and each candidate tweet, and a similarity threshold determines the classification~\cite{cer2018universal}.

 \item\textbf{TF-IDF + Cosine Similarity}:  
A traditional lexical matching method that encodes tweets using TF-IDF vectors with unigrams, bigrams, and trigrams. The averaged TF-IDF vector of a user’s previous tweets forms the author profile, and cosine similarity is used to compare it to test tweets~\cite{stamatatos2009survey}.

 \item\textbf{GZip Compression Distance (NCD)}:  
This compression-based baseline constructs a profile by concatenating all of a user’s previous tweets. It computes the Normalized Compression Distance (NCD) between the profile and each test tweet using GZip. A dynamic threshold based on median distances is used for classification~\cite{potha2017improved}.

 \item\textbf{Siamese Network with GloVe + LSTM}:  
This neural architecture uses pretrained GloVe embeddings and an LSTM encoder to represent tweets. A Siamese network computes the cosine similarity between a pair of encoded tweets, and the result is passed through a sigmoid output layer to determine authorship likelihood~\cite{boenninghoff2019similarity}.

 \item \textbf{Siamese Network with SBERT Embeddings}:  
Each tweet is encoded using a pretrained SBERT model to capture semantic meaning. The Siamese architecture computes cosine similarity between tweet embeddings and uses a dense layer to classify whether the tweets belong to the same author~\cite{reimers2019sentence}.\\
\end{itemize}

\subsection*{LLMs Knowledge Cut-off Dates} The knowledge cut-off dates of the evaluated LLMs for assessing their generalization performance on ``unseen data" (Section~\ref{subsec:authorship_verification} and ~\ref{sec:exp authorship}) are presented in Table~\ref{tab:model_cutoffs}. As it can be observed, all dates are before January 2024 onward, the date for which we we collected the new (``unseen data"). 

\begin{table}[htbp]
\centering
\small
\caption{LLMs and their knowledge cutoff dates }
\label{tab:model_cutoffs}
\begin{tabular}{ll}
\toprule
\textbf{Model} & \textbf{Cut-off Date} \\ 
\midrule
GPT-4 & Oct. 2023 \\
GPT-4o & Sep. 2023 \\
GPT-3.5-Turbo & Aug. 2021 \\
Gemini 1.5 Pro & Nov. 2023 \\
DeepSeek-V3 & Oct. 2023 \\
Llama 3.2 & Dec. 2023 \\
\bottomrule
\end{tabular}
\end{table}

\subsection*{Impact of Sampling Strategies}


The values reported in Table~\ref{tab:sampling_effects} for the User Effect and Tweet Effect were computed as follows:

\begin{itemize}
    \item \textbf{User Effect}: For each negative post sampling strategy, we calculated the maximum and minimum accuracy across the three user post sampling strategies (Rnd, Rec, Top), and averaged these differences across all negative post sampling strategies:
    \[
    U_{\text{effect}} = \frac{1}{|T|}\sum_{t \in T}\left(\max_{u \in U}(A_{t,u}) - \min_{u \in U}(A_{t,u})\right)
    \]

    \item \textbf{Tweet Effect}: For each positive user sampling strategy, we calculated the maximum and minimum accuracy across the five negative post sampling strategies (Reciprocal, Followees-only, Followers-only, Random, Topic-similar), and averaged these differences across all positive user sampling strategies:
    \[
    T_{\text{effect}} = \frac{1}{|U|}\sum_{u \in U}\left(\max_{t \in T}(A_{t,u}) - \min_{t \in T}(A_{t,u})\right)
    \]
\end{itemize}

Where:
\begin{itemize}
    \item \( T \) represents the set of negative post sampling strategies: Reciprocal, Followees-only, Followers-only, Random, Topic-similar.
    \item \( U \) represents the set of positive user sampling strategies: Random (Rnd), Recently active (Rec), Top-active (Top).
    \item \( A_{t,u} \) denotes the accuracy obtained for a specific negative post sampling strategy \( t \) and positive user sampling strategy \( u \).
\end{itemize}

Average performance analysis (Figure.~\ref{fig:classwise_f1_comparison} and Table~\ref{tab:f1_centroids}) indicates that GPT--4 consistently leads, delivering high and balanced performance across both Negative Evaluation Posts and Positive Evaluation Posts (genuine user's tweets) classes. Gemini and DeepSeek represent a robust second tier, achieving strong accuracy but with a clearer bias toward the Positive class. While GPT--3.5--Turbo stands out as the most balanced model, it does so at a noticeably lower overall performance level. Traditional models like Random Forest exhibit substantial Positive-class bias, and other baselines such as BERT and Llama highlight significant performance gaps and limitations. Similarly, classical baseline methods (e.g., USE, TF-IDF, Gzip, and SIAMESE variants) show varying degrees of accuracy and imbalance.

\begin{figure*}[htbp]
    \centering
    \includegraphics[width=0.7\textwidth]{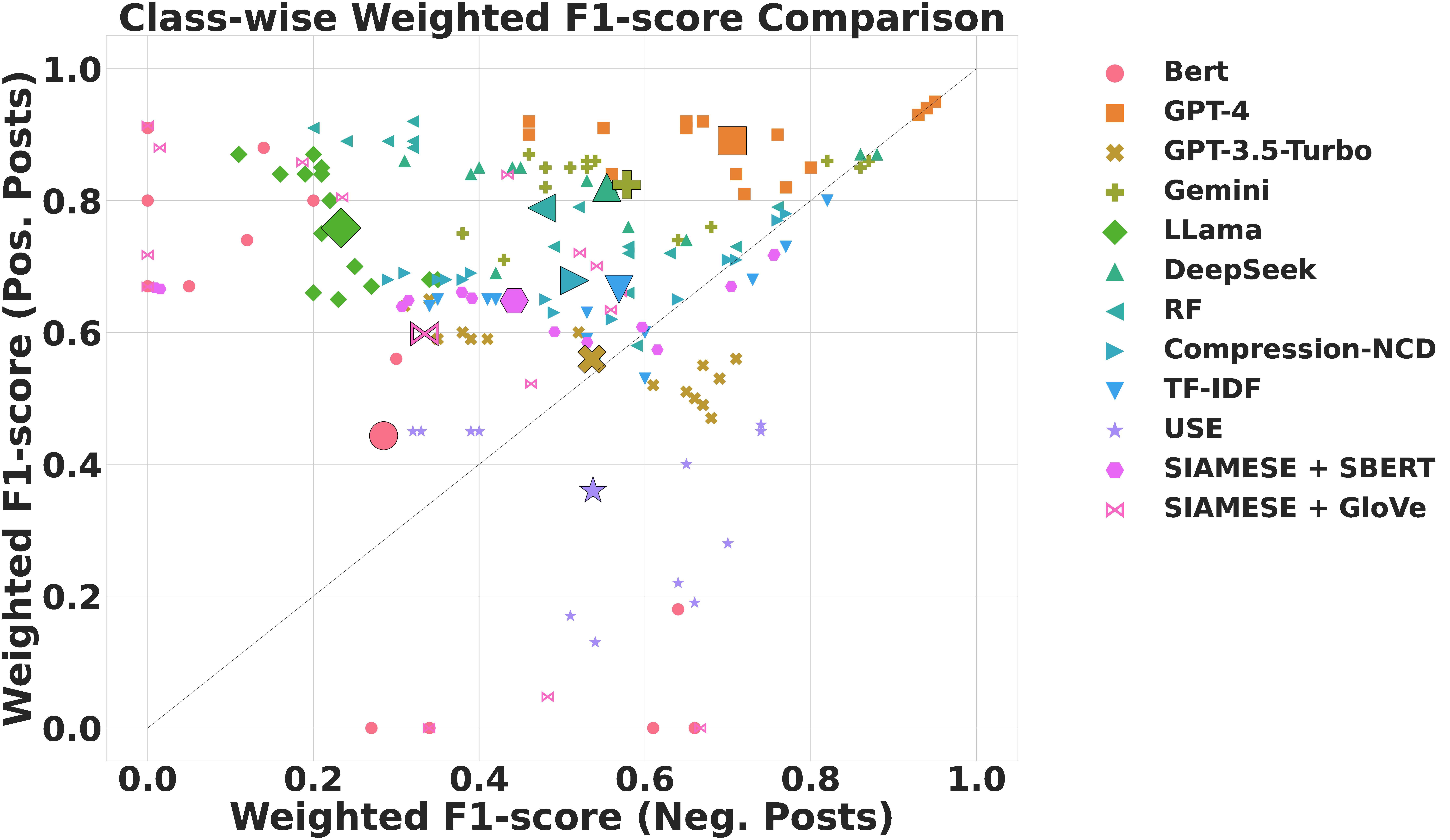}
    \caption{Mean class-wise weighted F1-scores for each model. Points above the dashed line ($y{=}x$) indicate better performance on Positive Evaluation Posts; points below indicate better performance on Negative Evaluation posts.}
    \label{fig:classwise_f1_comparison}
\end{figure*}

\begin{table}[htbp]
  \centering
  \small
  \sisetup{
    table-format = 1.2,
    detect-weight = true,
    detect-inline-weight = math
  }
    \caption{Mean for weighted  F1‑scores per class and absolute balance gap.}
  \label{tab:f1_centroids}
  \begin{tabular}{l
                  S[table-format=1.2]
                  S[table-format=1.2]
                  S[table-format=1.2]}
    \toprule
    \textbf{Model} &
    {\textbf{F1 (Neg.)}} &
    {\textbf{F1 (Pos.)}} &
    {\textbf{Gap}} \\
    \midrule
    \textbf{GPT‑4}           & \bfseries 71 & \bfseries 89 & 18 \\
    Gemini 1.5 Pro           & 58 & 82 & 24 \\
    DeepSeek                 & 55 & 82 & 27 \\
    GPT‑3.5‑Turbo            & 54 & 56 & \bfseries 2 \\
    RF                       & 48 & 79 & 31 \\
    Llama 3.2                & 23 & 76 & 53 \\
    BERT                     & 29 & 44 & 15 \\
    USE                      & 54 & 36 & 18 \\
    TF-IDF                   & 57 & 66 & 9 \\
    Gzip                     & 52 & 68 & 16 \\
    SIAMESE + SBERT          & 44 & 65 & 21 \\
    SIAMESE + GloVe          & 33 & 60 & 27 \\
    \bottomrule
  \end{tabular}
\end{table}

Figure~\ref{fig:precision_recall_comparison} details the class-specific precision and recall for the authorship verification task. GPT--4 maintains a high precision (0.803) for Class 0, effectively reducing false positives, alongside an excellent recall (0.939) for Class 1, correctly identifying most user-authored tweets. Gemini and DeepSeek show strong balanced recall (approximately 0.87 and 0.86) for Class 1, but lower precision (0.66 and 0.62) for Class 0, reflecting moderate false-positive rates. While GPT--3.5--Turbo achieves high recall (0.930) for Class 0, its precision is notably low (0.391), indicating aggressive labeling with many false positives; inversely, it shows high precision (0.931) but low recall (0.402) for Class 1. Traditional methods such as Random Forest, TF-IDF, Gzip, SIAMESE+SBERT, SIAMESE+GloVe, and USE exhibit varied levels of performance, with notable biases and moderate precision-recall trade-offs. In contrast, BERT and Llama demonstrate significantly limited performance across both precision and recall metrics. Overall, these precision-recall analyses reinforce GPT--4's robustness and superior capability in accurately verifying authorship with balanced performance.

\begin{figure*}[htbp]
    \centering
    \includegraphics[width=.8\linewidth]{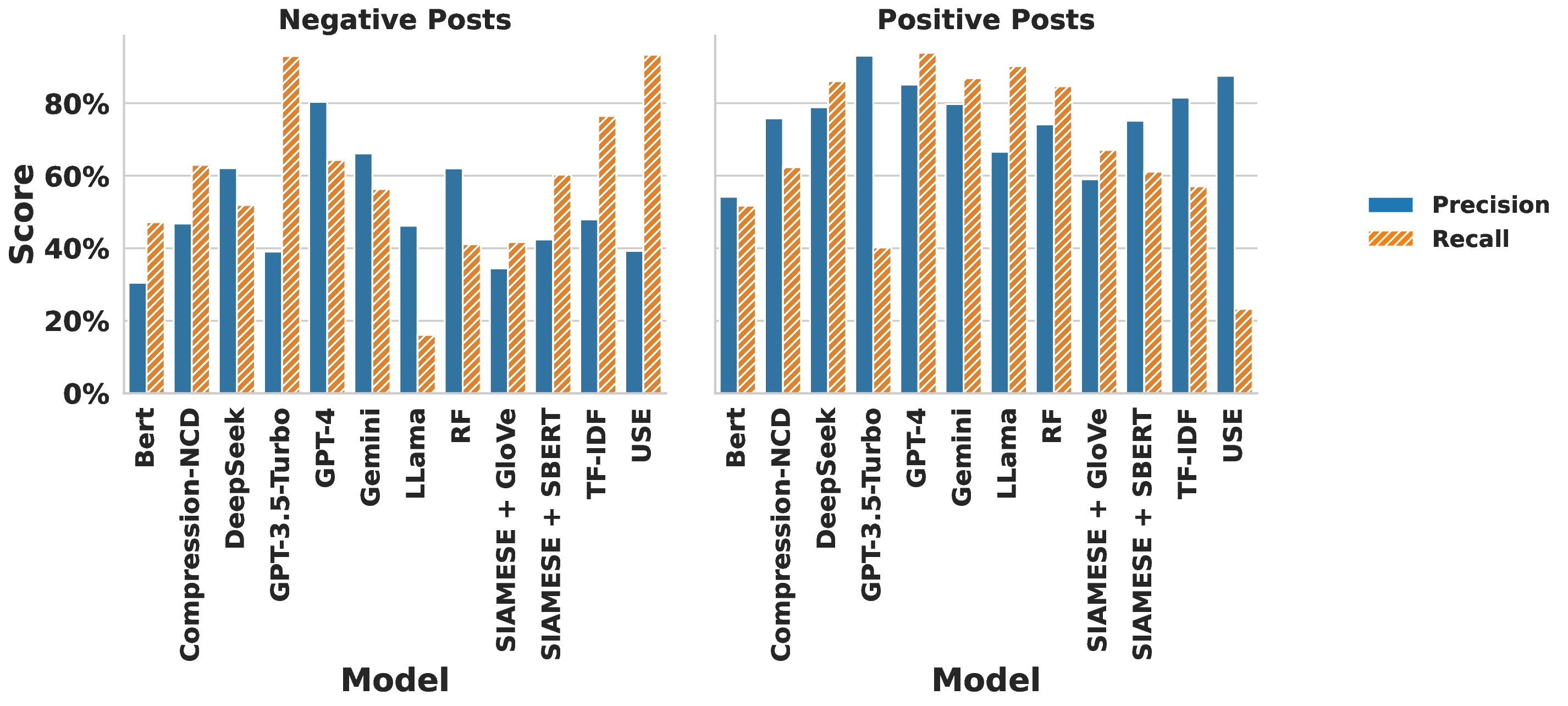}
    \caption{Precision and recall comparison across models for the authorship verification task.}
    \label{fig:precision_recall_comparison}
\end{figure*}

Figure~\ref{fig:metrics_heatmap} summarizes the mean evaluation metrics (accuracy, precision, recall, and F1-score) across evaluated models for the authorship verification task. GPT--4 achieves the highest overall performance, with accuracy (85.08\%) and F1-score (0.8898), reflecting a robust balance between precision (85.16\%) and recall (93.82\%). Gemini and DeepSeek exhibit strong and balanced results, with Gemini slightly outperforming DeepSeek in accuracy (76.25\%) and F1-score (0.8240). Conversely, GPT-3.5-Turbo demonstrates an imbalanced performance, with exceptional precision (93.17\%) but notably low recall (40.11\%), indicative of conservative classification behavior. Traditional methods like Random Forest provide moderate and balanced performance (accuracy: 73.25\%), whereas baseline methods such as Llama 3.2 (accuracy: 64.41\%, high recall but lower precision) and BERT (accuracy: 51.44\%) show clear limitations. Additional classical baselines (USE, TF-IDF, Gzip, SIAMESE+SBERT, and SIAMESE+GloVe) also exhibit varied levels of accuracy and balance, underscoring GPT--4's superior performance and robustness in authorship verification.

\begin{figure}[htbp]
    \centering
    \includegraphics[width=\linewidth]{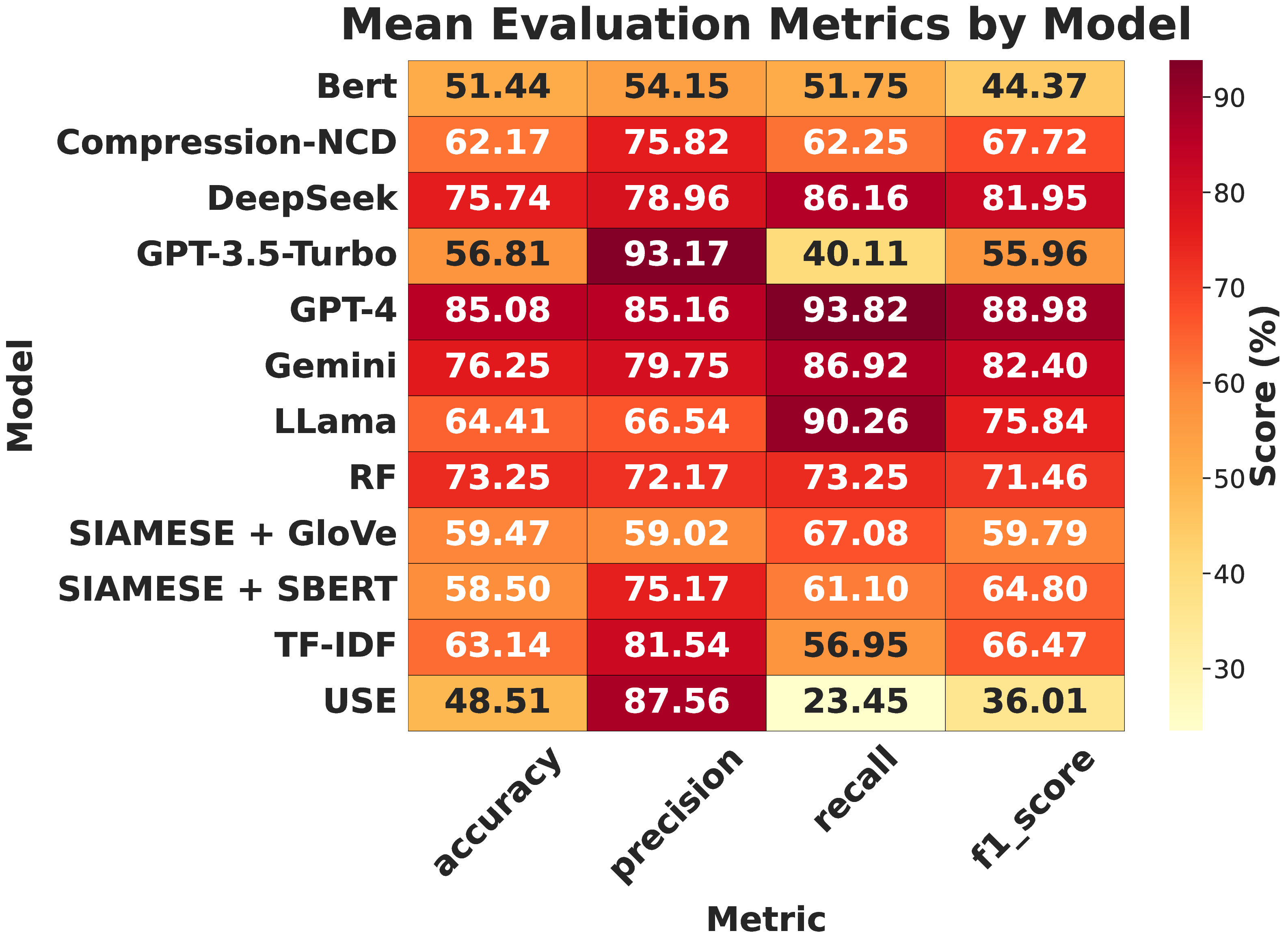}
    \caption{Heatmap illustrating mean evaluation metrics (accuracy, precision, recall, and F1-score) across different models evaluated on the authorship verification task. Higher scores (darker colors) indicate stronger performance.}
    \label{fig:metrics_heatmap}
\end{figure}

\section{Social Media Post Generation}
\label{sec:appendix_D}

\subsection*{Model Access and Hyperparameters} For tweet generation, we accessed OpenAI's GPT-4o using a batch API to reduce cost by submitting prompts in bulk. In contrast, Gemini, DeepSeek, and Llama (via Ollama) were accessed through individual non-batched calls. The generation temperature was uniformly set to 0.7 across all models to promote content diversity while avoiding degenerate or repetitive outputs.

\subsection*{Descriptions of Post Generation Baselines} To complement our evaluation of LLMs on the social media post generation task, we implemented a set of traditional baseline models that reflect diverse approaches to content generation. Below, we provide a brief description of each method.

\begin{itemize}
 
  \item  \textbf{Markov Chain}: A classical probabilistic model that generates new tweets by modeling transition probabilities between word sequences. We train a second-order Markov chain using a user's prompt tweets and sample new sentences constrained to tweet length. This method captures frequent local patterns but lacks global coherence or semantic understanding.~\cite{shannon1948mathematical,freitas2015reverse}

  \item  \textbf{BART-large}: A denoising autoencoder for sequence-to-sequence generation that uses a bidirectional encoder and a left-to-right decoder. Pretrained on large-scale corpora with text corruption tasks, BART is well-suited for conditional generation tasks, including style imitation and text rewriting.~\cite{lewis2019bart}

  \item \textbf{T5-large}: A unified text-to-text transformer that reformulates all NLP tasks—including generation—as text-to-text problems. The \texttt{large} variant is pretrained on the C4 corpus with a multi-task objective and fine-tuned for conditional generation.~\cite{raffel2020exploring}

\subsubsection*{Qualitative Analysis for Task II (Case Studies)}
Table~\ref{tab:scioly_llm_coverage} compares tweets originally posted by one user to tweets generated by four LLMs (DeepSeek, GPT-4o, Gemini, and Llama). The table highlights two main topics from the user's tweets: a\textit{ppreciation for volunteers} and \textit{recognition of alumni scholarship achievements}. Among the LLM models, DeepSeek and GPT-4o deliver the closest matches. Both accurately reflect the volunteer appreciation theme with similar enthusiasm, although GPT-4o slightly lacks specifics regarding the alumni scholarship. Gemini   effectively praises volunteers but becomes too general when mentioning scholarships. Lastly, Llama  maintains relevance to Science Olympiad but often overlooks specific details, shifting attention to general team accomplishments and different scholarships. Overall, DeepSeek and GPT-4o provide tweets most faithful to the user's original content, while Gemini and Llama are thematically relevant but less precise.

\begin{table}[htp]
\centering
\caption{\small Comparison of one user's tweets and LLM-generated tweets.}
\label{tab:scioly_llm_coverage}
\small
\renewcommand{\arraystretch}{1.35}
\begin{tabularx}{\columnwidth}{X}
\toprule

\textbf{Theme A – Volunteer appreciation} \\[2pt]
\highlightcell\textbf{Real:} @MDSciOly Congrats to Russ for being recognized for his incredible Science Olympiad volunteer work! We are so proud! @SOAlumniNetwork \\[2pt]

\textbf{Gen (DeepSeek):} Shoutout to all the incredible coaches and volunteers who make Science Olympiad possible. Your passion fuels the next generation of innovators! \#ThankYou \\
\textit{Comment:} Matches closely; same focus on volunteers with similar enthusiasm. \\[2pt]

\textbf{Gen (GPT-4o):} Shoutout to all the incredible volunteers who make Science Olympiad possible! Your dedication inspires future scientists every day. Thank you! \#VolunteerAppreciation \\
\textit{Comment:} Excellent match; closely mirrors topic, appreciation tone, and enthusiasm. \\[2pt]

\textbf{Gen (Gemini):} Volunteers are the heart of Science Olympiad! Thank you for your dedication. \#volunteerappreciation \\
\textit{Comment:} Good match on volunteers, but a bit general compared to the original. \\[2pt]

\textbf{Gen (Llama):} Meet our incredible teams of 7,800+ students from across 50 states gearing up for the ultimate STEM competition - Science Olympiad! \\
\textit{Comment:} Stays related to Science Olympiad but misses the volunteer focus. \\

\midrule

\textbf{Theme B – Alumni success \& scholarships} \\[2pt]
\highlightcell\textbf{Real:} This Science Olympiad alum and valedictorian candidate just won the prestigious \$40,000 Edison Scholarship! Tanya is MIT-bound and plans to major in chemical engineering. Way to go! \#STEM \#scienceolympiad \\[2pt]

\textbf{Gen (DeepSeek):} Meet Tanya, a Science Olympiad alum and valedictorian who just won the \$40K Edison Scholarship! Your future starts here. \#STEMSuccess \\
\textit{Comment:} Good match; includes the same name, scholarship amount, and excitement. \\[2pt]

\textbf{Gen (GPT-4o):} Big congrats to our alumni who are making waves in the STEM world! Your achievements showcase the lasting impact of Science Olympiad. Keep shining! \#ScienceOlympiadAlumni \\
\textit{Comment:} Good topical alignment about alumni success, though it omits specific scholarship details. \\[2pt]

\textbf{Gen (Llama):} Congratulations to our Division C winners who each won \$1,000 college scholarships from DuPont! Your hard work pays off! \\
\textit{Comment:} On-topic with scholarships but uses different details and lacks personal specifics. \\[2pt]

\textbf{Gen (Gemini):} Science Olympiad alumni, where are you now? Share your success stories! \#sciolyalumni \\
\textit{Comment:} Mentions alumni generally but misses details about the specific scholarship and winner. \\

\bottomrule
\end{tabularx}
\end{table}

\end{itemize}

\section{User Study}
\label{sec:appendix_E}

Table~\ref{tab:user_survey} summarises the 60 ratings collected for each model.  For every system we report three descriptive statistics: the mean score on the five‑point \emph{“definitely not me” (1) – “definitely me” (5)} scale, its associated 95\% confidence interval (CI), and the \emph{Top‑2‑box} proportion—i.e., the share of responses falling in the two most positive categories, an intuitive acceptance rate.

Gemini and Llama obtain the strongest reception: almost three‑quarters of their tweets are marked \emph{probably me} or \emph{definitely me}, and both achieve a mean rating of 3.95/5.  DeepSeek and GPT‑4o lag by roughly 0.30 points in mean score and by about 5–8 percentage points in the positive bins metric, implying a modest—but not dramatic—loss of perceived author likeness.  Because the 95 \% CIs overlap across all four systems, these differences should be viewed as suggestive rather than conclusive.

To test whether the full five‑category rating distributions differ by model, we conducted a Pearson $\chi^{2}$ test on the complete $4\times5$ contingency table (four models, five response categories).  The result, $\chi^{2}=7.29$ with 12 degrees of freedom ($p=0.84$), fails to reject the null hypothesis of identical distributions.  In practical terms, the available 60 judgments per model do not provide sufficient power to claim a statistically significant winner, even though Gemini and Llama trend higher on both descriptive metrics.

\begin{table}[ht]
  \caption{Participant‑likeness ratings on a 1–5 scale (higher = more authentic).}
  \label{tab:user_survey}
\centering
\small
\begin{tabular}{lcc}
\toprule
\textbf{Model} & \textbf{Mean $\pm$ 95\% CI} & \textbf{Positive Bins \%} \\
\midrule
DeepSeek        & $3.68 \pm 0.34$ & 65.0 \\
Gemini  & $\mathbf{3.95 \pm 0.32}$ & \textbf{73.3} \\
GPT-4o          & $3.67 \pm 0.36$ & 68.3 \\
Llama      & $\mathbf{3.95 \pm 0.29}$ & \textbf{73.3} \\
\bottomrule
\end{tabular}
\end{table}

 Table~\ref{tab:user_demographics} presents detailed demographic characteristics and additional information about the user study participants.

\begin{table}[htp]
\centering
\caption{Demographics of User Study Participants}
\label{tab:user_demographics}
\small
\renewcommand{\arraystretch}{1.2}

\begin{tabularx}{0.5\textwidth}{|>{\bfseries\centering\arraybackslash}m{0.18\textwidth}|X|}
\hline
\makecell{Age Range} &
\begin{itemize}[nosep,leftmargin=*]
    \item \textbf{[22--26)}: 63.64\%
    \item \textbf{[26--30]}: 36.36\%
\end{itemize}\\ \hline

\makecell{Gender} &
\begin{itemize}[nosep,leftmargin=*]
    \item \textbf{Male}: 91.67\%
    \item \textbf{Female}: 8.33\%
\end{itemize}\\ \hline

\makecell{Tweet Activity} &
\begin{itemize}[nosep,leftmargin=*]
    \item \textbf{Daily}: 66.67\%
    \item \textbf{A few times/week}: 25.00\%
    \item \textbf{Weekly}: 8.33\%
\end{itemize}\\ \hline

\makecell{Main Tweet Topics} &
\begin{itemize}[nosep,leftmargin=*]
    \item \textbf{Entertainment}: 91.67\%
    \item \textbf{Technology/AI}: 75.00\%
    \item \textbf{Sports}: 58.33\%
    \item \textbf{Science}: 50.00\%
    \item \textbf{Politics}: 41.67\%
    \item \textbf{Personal updates}: 33.33\%
    \item \textbf{E-commerce}: 33.33\%
    \item \textbf{Religion}: 8.33\%
\end{itemize}\\ \hline

\end{tabularx}
\end{table}

\subsection*{Qualitative Analysis for User Study (Case Studies)}
Table~\ref{tab:heri_llm_coverage} compares original tweets from one participant with tweets generated by four different LLMs (GPT-4o, DeepSeek, Gemini, and Llama), focusing on two specific topics: \textit{fintech skepticism} and \textit{car purchasing decisions}. Among these models, GPT-4o best captures the user's style, accurately reflecting the original tweets' topics and conversational tone, though it sometimes sounds slightly more formal. DeepSeek does very well in fintech discussions, matching the user's viewpoint closely, but struggles with other topics like cars, where it drifts into unrelated general tech content. Gemini often misses the user's specific interests entirely, defaulting instead to generic technology-related statements. Lastly, Llama picks up on contextual details (e.g., Canadian outdoors) but frequently misses the user's main points about fintech and specific vehicle choices, shifting instead toward broader, less relevant subjects.

\begin{table}[!t]
\centering
\caption{\small Comparison of one participant's original tweets and LLM-generated tweets.}
\label{tab:heri_llm_coverage}
\small
\renewcommand{\arraystretch}{1.35}
\begin{tabularx}{\columnwidth}{X}
\toprule

\textbf{Theme A – Fintech skepticism} \\[1pt]

\textbf{An overview of the user:} 
Male, 26 years old, Software Developer (USA). Tweets weekly, mainly about Technology/AI and Entertainment. \\[4pt]

\highlightcell \textbf{Real sample tweet:} Are there any cons to @Wealthsimple? So far I don't see why anyone would use a traditional bank. \\[1pt]

\textbf{Gen (DeepSeek):} @Wealthsimple Your app is a game-changer. Traditional banks should be worried. \\
\textit{Comment:} Closely matches topic and tone; clearly shares the user's viewpoint. \\
\textit{User's survey response:} Definitely me\\[1pt]

\textbf{Gen (GPT-4o):} Let's talk smart investments. With the rise of fintech platforms, do you think traditional banks will become obsolete? \#Fintech \#Investing \\
\textit{Comment:} Keeps the casual question-and-answer style and fintech focus, but slightly more formal. \\
\textit{User's survey response:} Definitely me\\[1pt]

\textbf{Gen (Gemini):} Is it just me, or is AI getting smarter every day? Kind of exciting, kind of terrifying. \#AI \\
\textit{Comment:} Completely off-topic; shifts to general AI instead of finance. \\
\textit{User's survey response:} Definitely me \\[1pt]

\textbf{Gen (Llama):} Thinking of switching to a Linux distribution for my next project – any recommendations? \#Linux \\
\textit{Comment:} Misses the financial topic completely; instead emphasizes a general tech persona. \\
\textit{User's survey response:} Definitely not me \\

\midrule

\textbf{Theme B – Car decisions} \\[2pt]
\highlightcell \textbf{Real:} First time Subaru is stuck. Hmm should I get a Lexus GX or a Land Cruiser? \\

\textbf{Gen (GPT-4o):} Thinking about switching from Subaru to Lexus GX. Need something more rugged for those off-road adventures. Any thoughts? \#CarTalk \\
\textit{Comment:} Excellent match; closely follows user's original question and specific vehicle brands. \\
\textit{User's survey response:} Probably me\\[1pt]

\textbf{Gen (Llama):} Planning a road trip to explore the Canadian wilderness – any recommendations for off-road gear? \#offroadadventures \#Canada\\
\textit{Comment:} Captures outdoor theme but doesn't reflect user's original car-brand comparison. \\
\textit{User's survey response:} Unsure\\[1pt]

\textbf{Gen (Gemini):} Electric cars are the future. Just sayin’. \#EV \#electriccar\\
\textit{Comment:} Briefly touches on cars, but too general; misses user's interest in specific brands and off-road vehicles. \\
\textit{User's survey response:} Probably me\\[1pt]

\textbf{Gen (DeepSeek):} Just pre-ordered the new Framework laptop. Modular design is the future. \#RightToRepair \\
\textit{Comment:} Completely unrelated topic; shifts to a generic tech subject rather than cars. \\
\textit{User's survey response:} Probably not me\\

\bottomrule

\end{tabularx}
\end{table}

\section{User Attribute Inference}
\label{sec:appendix_F}

\subsection*{Misclassification Patterns}
Table~\ref{tab:task3_confusion} summarizes the most frequent true–predicted label confusions in Task III, along with their occurrence counts. Across both interest and occupation inference, most errors occur between semantically adjacent categories with overlapping topical or stylistic signals, rather than between unrelated domains. This suggests that misclassifications largely reflect boundary ambiguity in social media language rather than systematic model failure.


\begin{table}[t]
\small
\centering
\caption{Most Frequent Confusion Pairs in User Attribute Inference (Task III)}
\label{tab:task3_confusion}
\setlength{\tabcolsep}{5pt}
\renewcommand{\arraystretch}{1.1}
\begin{tabularx}{\columnwidth}{>{\centering\arraybackslash}p{2.1cm} X X c}
\hline
\textbf{Category Type} & \textbf{True Category} & \textbf{Predicted Category} & \textbf{Count} \\
\hline
\multirow{2}{*}{\raisebox{-0.5\height}{Interest (IAB)}}
& Entertainment & Pop Culture & 45 \\
& Pop Culture & Entertainment & 18 \\
\hline
\multirow{2}{*}{\raisebox{-4.5\height}{Occupation ($L_1$)}}
& Education, Training, and Library
& Arts, Design, Entertainment, Sports, and Media & 33 \\
& Management
& Arts, Design, Entertainment, Sports, and Media & 22 \\
\hline
\multirow{2}{*}{\raisebox{-4.5\height}{Occupation ($L_2$)}}
& Entertainers and Performers
& Media and Communication Workers & 43 \\
& Media and Communication Workers
& Entertainers and Performers & 41 \\
\hline
\end{tabularx}
\end{table}

\subsection*{Model Access and Hyperparameters} For the user attribute inference task, we set the generation temperature to \texttt{0} for all models to ensure consistent outputs and reproducibility. GPT-4o was accessed via OpenAI's \texttt{ChatCompletion} endpoint, Gemini through Google's \texttt{GenerativeModel} API, DeepSeek using its REST interface, and Llama via local inference through Ollama. All models were queried individually in non-batch mode, and responses were parsed to extract structured JSON-formatted predictions.

\begin{table}[h]
\centering
\caption{Interest Categories Based on IAB Content Taxonomy v3.1}
\label{tab:interest_categories}
\begin{tabular}{ll}
\hline
\textbf{ID} & \textbf{Interest Category} \\
\hline
I1  & Attractions \\
I2  & Automotive \\
I3  & Books and Literature \\
I4  & Business and Finance \\
I5  & Careers \\
I6  & Communication \\
I7  & Crime \\
I8  & Disasters \\
I9  & Education \\
I10 & Entertainment \\
I11 & Fine Art \\
I12 & Food \& Drink \\
I13 & Hobbies \& Interests \\
I14 & Home \& Garden \\
I15 & Law \\
I16 & Medical Health \\
I17 & Pets \\
I18 & Politics \\
I19 & Pop Culture \\
I20 & Science \\
I21 & Sports \\
I22 & Style \& Fashion \\
I23 & Technology \& Computing \\
I24 & Travel \\
I25 & Video Gaming \\
\hline
\end{tabular}
\end{table}

\begin{table}[t]
\small
\centering
\caption{Occupational Categories (Level 1) Based on SOC 2018}
\label{tab:occupation_l1}
\setlength{\tabcolsep}{4pt}
\renewcommand{\arraystretch}{1.05}
\begin{tabularx}{\columnwidth}{lX}
\hline
\textbf{ID} & \textbf{Occupation Category (Level 1)} \\
\hline
L1-1  & Accommodation and Food Services \\
L1-2  & Arts, Design, Entertainment, Sports, and Media Occupations \\
L1-3  & Community and Social Service Occupations \\
L1-4  & Computer and Mathematical Occupations \\
L1-5  & Education, Training, and Library Occupations \\
L1-6  & Healthcare Practitioners and Technical Occupations \\
L1-7  & Legal Occupations \\
L1-8  & Life, Physical, and Social Science Occupations \\
L1-9  & Management Occupations \\
L1-10 & Management, Business, Science, and Arts Occupations \\
L1-11 & Management, Business, and Financial Occupations \\
L1-12 & Office and Administrative Support Occupations \\
L1-13 & Production Occupations \\
L1-14 & Professional and Related Occupations \\
L1-15 & Professional, Scientific, and Technical Services \\
L1-16 & Protective Service Occupations \\
L1-17 & Sales and Office Occupations \\
L1-18 & Sales and Related Occupations \\
\hline
\end{tabularx}
\end{table}

\begin{table}[t]
\small
\centering
\caption{Occupational Categories (Level 2) Based on SOC 2018}
\label{tab:occupation_l2}
\setlength{\tabcolsep}{4pt}
\renewcommand{\arraystretch}{1.03}
\begin{tabularx}{\columnwidth}{lX}
\hline
\textbf{ID} & \textbf{Occupation Category (Level 2)} \\
\hline
L2-1  & Accommodation \\
L2-2  & Advertising, Marketing, Promotions, Public Relations, and Sales Managers \\
L2-3  & Arts and Design Workers \\
L2-4  & Assemblers and Fabricators \\
L2-5  & Business and Financial Operations Occupations \\
L2-6  & Computer Occupations \\
L2-7  & Computer and Information Systems Managers \\
L2-8  & Computer and Mathematical Occupations \\
L2-9  & Counselors, Social Workers, and Other Community and Social Service Specialists \\
L2-10 & Educational Instruction and Library Occupations \\
L2-11 & Entertainers and Performers, Sports and Related Workers \\
L2-12 & Financial Clerks \\
L2-13 & Firefighting and Prevention Workers \\
L2-14 & Health Technologists and Technicians \\
L2-15 & Information and Record Clerks \\
L2-16 & Law Enforcement Workers \\
L2-17 & Lawyers, Judges, and Related Workers \\
L2-18 & Legal Occupations \\
L2-19 & Librarians, Curators, and Archivists \\
L2-20 & Life Scientists \\
L2-21 & Life, Physical, and Social Science Occupations \\
L2-22 & Management Occupations \\
L2-23 & Mathematical Science Occupations \\
L2-24 & Media and Communication Occupations \\
L2-25 & Media and Communication Workers \\
L2-26 & Office and Administrative Support Occupations \\
L2-27 & Other Management Occupations \\
L2-28 & Other Protective Service Workers \\
L2-29 & Physical Scientists \\
L2-30 & Postsecondary Teachers \\
L2-31 & Professional, Scientific, and Technical Services \\
L2-32 & Retail Sales Workers \\
L2-33 & Sales Representatives, Services \\
L2-34 & Sales Representatives, Wholesale and Manufacturing \\
L2-35 & Scientific Research and Development Services \\
L2-36 & Social Science Occupations \\
L2-37 & Social Scientists and Related Workers \\
L2-38 & Top Executives \\
\hline
\end{tabularx}
\end{table}

\subsection*{Descriptions of User Attribute Inference Baselines} To complement our evaluation of LLMs on the user attribute inference task, we implemented a set of traditional baseline models. Below, we briefly describe each method along with their references.

\begin{itemize}

  \item\textbf{(Preoţiuc-Pietro et al. 2015)}: A probabilistic classifier that employs Word2Vec embeddings to represent tweets and spectral clustering to group semantically related words into clusters. It handles class imbalance through random oversampling, followed by classification using a Gaussian Process with an Automatic Relevance Determination (ARD) kernel.
  \item
\textbf{\cite{lewis2019bart}}: A fine-tuning approach leveraging a pre-trained BART model (\texttt{bart-large-mnli}) on a small subset of labeled user tweets (20\%). It employs random oversampling for balancing classes, evaluating performance on an independent test set.

  \item\textbf{\cite{michelson2010discovering}}: This method extracts named entities from tweets and maps them to DBpedia categories to build user interest profiles. It infers user attributes based on the frequency of categories associated with these entities, capturing topical interests rather than textual semantics.

  \item \textbf{\cite{pennacchiotti2011machine}}: A gradient boosting classifier that utilizes textual features extracted via TF-IDF vectorization of user tweets. The classifier addresses class imbalance using sample weighting to optimize predictive performance.

\end{itemize}

\subsection*{Qualitative Analysis for Task III (Case Studies)}

Table~\ref{tab:occupation_inference_EM_case} shows occupation predictions by four LLMs (Gemini, GPT-4o, DeepSeek, and Llama) based solely on tweets from a user involved in emergency management. Among the models, Gemini correctly identifies the occupation as ``Other Protective Service Workers," accurately capturing specialized professional cues such as ``Certified Emergency Manager," mentions of FEMA, and emergency response drills. In contrast, GPT-4o incorrectly emphasizes general community support themes and hashtags, leading to a prediction in social and community services. DeepSeek misinterprets hazard-related language as indicating frontline firefighting roles, failing to distinguish between emergency response coordination and direct hazard management. Finally, Llama provides an overly broad classification (``Public Service"), overlooking precise professional indicators and domain-specific terminology found clearly within the user's tweets.

\begin{table}[!t]
\centering
\caption{\small Occupation (L$_2$) inference from tweets.}
\label{tab:occupation_inference_EM_case}
\small
\renewcommand{\arraystretch}{1.35}
\begin{tabularx}{\columnwidth}{X}
\toprule
\textbf{User Bio (not shown to models):} Disaster technologist, inclement weather enthusiast, tender‑hearted public servant. Just trying to make the world a safer place. Views expressed are my own. \\[4pt]
\midrule
\textbf{Sample Tweets:} \\[2pt]
$\bullet$ ``I am an \textbf{internationally Certified Emergency Manager} with a Bachelor of Science in Emergency Management.''\\[2pt]
$\bullet$ ``I'm wrapping up the last bit of work‑fun for the week by reviewing the new \textbf{@FEMA ICS/NIMS courses}.''\\[2pt]
$\bullet$ ``Taught an \textbf{Incident Command System (ICS) course} with the brilliant @IDICworld today!''\\[2pt]
$\bullet$ ``Just wrapping‑up a fun day helping test our statewide response capability to a \textbf{complex coordinated cyber attack}.''\\[2pt]
$\bullet$ ``I can not say enough good things about the \textbf{Emergency Management Accreditation Program}! Urban or rural, private or public …''\\[2pt]
$\bullet$ ``Those are some pretty serious polygons! Stay safe friends! \#KSWX''\\[2pt]
$\bullet$ ``\#Preach! Everyone is entitled to \#SelfCare in any form, as long as it isn't hurting anyone else.''\\[4pt]
\midrule
\textbf{Ground‑Truth Occupation (SOC 2018):} Other Protective Service Workers (33‑9099) \\[2pt]
\midrule
\textbf{Model Predictions and Diagnostic Analysis:}\\[2pt]

\textbf{Gemini:} \emph{Other Protective Service Workers} \\
\textit{Comment:} Correct. The model accurately identified clear evidence, such as ``Certified Emergency Manager,'' ``ICS training,'' mentions of ``FEMA,'' and disaster-response drills, matching closely with the protective-service occupation.\\[4pt]

\textbf{GPT‑4o:} \emph{Counselors, Social Workers, and Other Community \& Social Service Specialists} \\
\textit{Comment:} Incorrect. The model was influenced heavily by general expressions of well-being and hashtags like \#SelfCare, overlooking specific professional terms related to emergency management.\\[4pt]

\textbf{DeepSeek:} \emph{Firefighting and Prevention Workers} \\
\textit{Comment:} Incorrect. The model overly focused on tweets about hazards and severe weather (e.g., ``polygons'' in weather warnings), mistakenly interpreting coordination and training roles as frontline firefighting.\\[4pt]

\textbf{Llama:} \emph{Public Service} \\
\textit{Comment:} Incorrect. Too broad and vague, this prediction missed the specific professional cues (e.g., certifications and specialized acronyms such as ``ICS,'' ``NIMS,'' and ``CEM'') clearly indicating emergency management work.\\
\bottomrule
\end{tabularx}
\end{table}

\subsection*{Interest and Occupation Category Taxonomies}
Tables~\ref{tab:interest_categories}, \ref{tab:occupation_l1}, and \ref{tab:occupation_l2} list the interest categories and the Level 1 and Level 2 occupation categories, respectively, as defined by the IAB Content Taxonomy and the 2018 SOC classification. These tables document the label space used for user attribute inference and support reproducibility of our experiments.

\clearpage

\section{Related Work}
\label{sec:appendix_G}

The research in social media analytics has been conducted using conventional machine learning and statistical methods~\cite{injadat2016data}. The introduction of transformer-based LLMs specially versatile ones such as Gemini and GPT, has marked a significant advancement in social media analytics. Next, we briefly review the recent studies that utilize LLMs in the three tasks of our interest.

Recent authorship attribution and verification studies increasingly use LLMs to capture distinctive writing styles from text alone~\cite{huertas2024understanding}. Fine-tuned transformer encoders (e.g., BERT, RoBERTa) applied to authorship tasks have achieved state-of-the-art accuracy, surpassing traditional stylometric approaches that rely on handcrafted features~\cite{hu2024instructav}. More recent works explore both fine-tuning and prompting strategies with advanced LLMs. For example, \citep{huang2024can} demonstrated that GPT-based LLMs can accurately verify authorship (and even attribute texts to the correct author among many candidates) in a zero-shot setting without task-specific training, essentially establishing new performance benchmarks. Other researchers have proposed prompt-based techniques to harness LLMs’ knowledge; for instance, a “PromptAV” method uses step-by-step stylometric cues to improve GPT-3.5’s verification accuracy and explainability, and a linguistically informed prompting approach similarly guides GPT-3.5/4 models to strong authorship verification results even without fine-tuning~\cite{hu2024instructav}. However, these LLM-driven approaches typically consider only textual content, omitting valuable contextual signals such as user profile bios or social network features. In the social media domain, ignoring such metadata can be limiting – social media posts are short and rife with slang, often making it difficult to identify the author from text alone~\cite{alsanoosy2024authorship}.  In contrast to prior studies, our work integrates \textbf{rich contextual metadata} (e.g., profile descriptions and network-derived features) into the authorship verification pipeline. We introduce  \textbf{systematic and robust user/post sampling strategies} to construct a diverse evaluation set, and we \textbf{mitigate potential data leakage} biases by using tweet content posted LLMs' knowledge cut-off.

A growing body of work uses LLMs to write (generate) social media posts, yet each study tackles a very specific goal. RePALM fine‑tuned GPT‑3.5 with a reinforcement‑learning reward that predicts likes and retweets, so it generates quote‑tweets optimized purely for popularity~\cite{yu2024repalm}. Pillai et al. prompted GPT‑4 to rewrite news headlines into tweets in three fixed persona styles (formal, casual, factual) to boost engagement \cite{pillai2025engagement}. Qiu et al. first predicted whether a user will retweet, quote or reply to a trending post and then let GPT‑4 craft the corresponding response, but the model still handled one interaction type at a time~\cite{qiu2025can}. Across these efforts, the LLM sees little more than the source post (plus an optional style tag); richer cues such as the author’s bio, follower network, or recent tweets are ignored—even though adding social signals during pre‑training is known to improve tweet representations \cite{zhang2023twhin,zhao2025amplifying}. Our study fills this gap by conditioning generation on a compact user‑context block—bio, follower/followee counts, and representative past tweets—and by rating the outputs on \textbf{four complementary dimensions}: (i) semantic fidelity (how closely each tweet’s meaning matches the user’s real posts), (ii) output diversity (coverage of topics and phrasings), (iii) stylistic congruence (faithfulness to the user’s voice or brand tone), and (iv) perceived authenticity (how natural and human‑like the tweets sound). More importantly, we assess LLMs’ post-generation capability by \textbf{asking real users to rate tweets} the models create from their own timelines.

Recent LLM-based approaches to social media user profiling typically target a single user attribute at a time – for example, classifying only a person’s occupation or their political leaning~\cite{Liu2024,wen2023towards}. To improve predictive power, many studies leverage auxiliary user data beyond the social media posts themselves. It is common to incorporate profile descriptions, full timelines, or social network cues along with the post text~\cite{Hong2021,wen2023towards}. For instance, users often self-report their job roles or hobbies in their  bios~\cite{Hong2021}, and some profiling methods feed such metadata (and even friendship information) into models alongside tweet content~\cite{wen2023towards}. Moreover, prior work seldom applies standard taxonomies for labeling user attributes. Instead, researchers usually define task-specific or coarse-grained categories – e.g. grouping occupations into a few broad classes~\cite{Liu2024} or using ad-hoc sets of interest topics – rather than mapping to established schemas. In contrast, our approach infers both occupation and personal-interest profiles simultaneously u\textbf{sing only each user’s tweet content}, without relying on any self-description or network features. We further constrain and explain the model’s outputs by grounding them in \textbf{official taxonomies} (the SOC for occupations and the IAB Tech Lab content taxonomy for interests), which enables more standardized, interpretable predictions in comparison to previous methods.


\end{document}